\documentclass[10pt,twocolumn,letterpaper]{article}

\usepackage[pagenumbers]{cvpr}

\usepackage[table,dvipsnames]{xcolor}

\definecolor{cvprblue}{rgb}{0.21,0.49,0.74}
\usepackage[pagebackref,breaklinks,colorlinks,citecolor=cvprblue]{hyperref}

\usepackage[ruled,vlined]{algorithm2e} 
\usepackage{amsbsy}
\usepackage{float}

\usepackage{newtxtext,newtxmath}
\usepackage{balance}

\usepackage[hyphens]{url}
\usepackage{graphicx} 
\usepackage{caption}

\definecolor{grey}{rgb}{0.5,0.5,0.5}

\usepackage{newfloat}
\usepackage{listings}
\usepackage{amsmath}
\usepackage{amssymb} 
\usepackage{algorithmicx}
\usepackage{booktabs}
\usepackage{multirow}
\usepackage{pifont}
\usepackage{microtype}

\setitemize{noitemsep,topsep=0pt,parsep=0pt,partopsep=0pt}

\newcommand{\cmark}{\ding{51}}

\def\paperID{****}

\def\confName{CVPR}
\def\confYear{2026}

\title{FMPose3D: monocular 3D pose estimation via flow matching}
\author{\textbf{Ti Wang, Xiaohang Yu, Mackenzie Weygandt Mathis} \\
École Polytechnique Fédérale de Lausanne (EPFL) \\\texttt{ \small mackenzie.mathis@epfl.ch
}
}

\begin{document}
\maketitle
\begin{abstract}
Monocular 3D pose estimation is fundamentally ill-posed due to depth ambiguity and occlusions, thereby motivating probabilistic methods that generate multiple plausible 3D pose hypotheses. In particular, diffusion-based models have recently demonstrated strong performance, but their iterative denoising process typically requires many timesteps for each prediction, making inference computationally expensive. In contrast, we leverage Flow Matching (FM) to learn a velocity field defined by an Ordinary Differential Equation (ODE), enabling efficient generation of 3D pose samples with only a few integration steps. We propose a novel generative pose estimation framework, FMPose3D, that formulates 3D pose estimation as a conditional distribution transport problem. It continuously transports samples from a standard Gaussian prior to the distribution of plausible 3D poses conditioned only on 2D inputs. Although ODE trajectories are deterministic, FMPose3D naturally generates various pose hypotheses by sampling different noise seeds. To obtain a single accurate prediction from those hypotheses, we further introduce a Reprojection-based Posterior Expectation Aggregation (RPEA) module, which approximates the Bayesian posterior expectation over 3D hypotheses. FMPose3D surpasses existing methods on the widely used human pose estimation benchmarks Human3.6M and MPI-INF-3DHP, and further achieves state-of-the-art performance on the 3D animal pose datasets Animal3D and CtrlAni3D, demonstrating strong performance across both 3D pose domains. The code is available at \href{https://github.com/AdaptiveMotorControlLab/FMPose3D}{https://github.com/AdaptiveMotorControlLab/FMPose3D}.
\vspace{-6pt}
\end{abstract}    
\vspace{-3pt}
\section{Introduction}

Monocular 3D pose estimation aims to recover 3D body joint locations from a single image or video. This task is fundamental to a wide range of applications, including computer animation, human–computer interaction, and action recognition. Most recent works adopt a 2D-to-3D lifting paradigm \cite{martinez2017simple,zhao2019semantic,li2019generating,zheng20213d,zhang2022mixste,shan2023diffusion} that decomposes the problem into two stages: (i) detecting 2D joint locations with off-the-shelf keypoint detectors \cite{chen2018cascaded,sun2019deep}, and (ii) lifting these 2D joints to their corresponding 3D coordinates. This formulation decouples the challenging 3D pose estimation problem from the more mature 2D keypoint detection stage, mitigating appearance and background variations and yielding stable inputs for 3D lifting.

Lifting a 2D pose to 3D from a single view is inherently ambiguous, as multiple 3D configurations can produce the same 2D projection.
This ambiguity makes deterministic models that learn a single mapping from 2D to 3D insufficient, as they tend to regress to an averaged prediction.
To mitigate this issue, some works \cite{jahangiri2017generating,li2019generating,sharma2019monocular,oikarinen2021graphmdn} adopt a multi-hypothesis strategy, producing a set of plausible 3D poses for each 2D input. This formulation allows the model to capture uncertainty and represent multiple valid interpretations of the same 2D pose. 
Early probabilistic works based on Mixture Density Network (MDN) \cite{li2019generating,oikarinen2021graphmdn} generate a fixed number of candidate poses with limited diversity. 
Recent diffusion-based approaches \cite{gong2023diffpose,holmquist2023diffpose,shan2023diffusion,wang2024text} reformulate 3D pose lifting as a reverse diffusion process that progressively denoises random noise conditioned on 2D inputs to sample diverse and 2D-consistent 3D poses. Despite their effectiveness, diffusion-based methods remain computationally expensive: their iterative denoising process requires many sequential sampling steps for each hypothesis, resulting in substantial inference overhead. Even with accelerated samplers such as Denoising Diffusion Implicit Models (DDIM) \cite{song2021denoising}, achieving strong accuracy typically still requires 10-15 steps. This unfavorable trade-off between accuracy and inference speed severely limits their applicability in real-time scenarios.

On the other hand, Flow Matching (FM) \cite{lipman2023flow,liu2023rectifiedflow} has recently emerged as a powerful class of generative models, capable of capturing complex data distributions while maintaining efficient and deterministic sampling. Rather than learning a stochastic denoising process governed by a Stochastic Differential Equation (SDE), FM learns a deterministic velocity field governed by an Ordinary Differential Equation (ODE), which continuously transports samples from a simple noise distribution to the target data distribution. In practice, Flow Matching enables fast and stable sampling, requiring only a few ODE integration steps, or even a single step with properly learned velocity fields \cite{wang2025editor,gui2025depthfm}.

Building upon this insight, we propose \textbf{FMPose3D}, a new paradigm for 3D pose estimation via flow matching.
Instead of treating pose lifting as a deterministic regression task, we formulate it as a conditional distribution transport problem, where the 2D pose serves as a guiding signal. The model learns a deterministic velocity field, governed by an ODE, that transports samples from a simple Gaussian prior to a distribution of plausible 3D poses conditioned on a 2D input. 
While each ODE trajectory is deterministic for a given noise seed and 2D input, sampling different seeds yields diverse 3D pose hypotheses for the same 2D input, enabling multi-hypothesis modeling within our framework.

To effectively distill these hypotheses into a single, more robust prediction, we further propose a novel \textbf{Reprojection-based Posterior Expectation Aggregation (RPEA)} module. RPEA aims to compute the posterior expectation of the 3D pose, which corresponds to the Bayes-optimal estimator under squared-error loss. Since the true posterior probability is intractable, we introduce a principled approximation: a plausible 3D pose must be consistent with its 2D observation. Therefore, we use the 2D re-projection error as an intuitive and effective proxy for the likelihood of each hypothesis. Aggregation can be performed joint-wise or pose-wise, and the resulting joints are assembled into the final 3D pose. Compared with uniform averaging across hypotheses or per-joint selection of the single best candidate, RPEA is theoretically closer to the Bayes-optimal solution and empirically yields superior accuracy.

Our contributions can be summarized as follows:
\begin{itemize}
\item We propose \textbf{FMPose3D}, a new generative framework that formulates 3D pose estimation as a conditional distribution transport problem, efficiently transporting Gaussian noise to plausible 3D poses conditioned on 2D inputs. To our knowledge, FMPose3D is the first work to successfully leverage flow matching for lifting 2D-to-3D poses.

\item We propose a new aggregation module, \textbf{RPEA}, which is motivated by Bayesian decision theory and fuses multiple 3D pose hypotheses into a single, robust prediction.

\item Extensive experiments on both human and non-human animal 3D pose datasets demonstrate the effectiveness of the proposed framework.

\end{itemize}

\section{Related Work}

\subsection{Monocular 3D Pose Estimation}

Recent 3D pose estimation works adopt a 2D-to-3D lifting pipeline \cite{martinez2017simple,zhao2019semantic,zheng20213d,zhang2022mixste,li2019generating,shan2023diffusion}.
Existing methods fall into two families: deterministic methods (single estimate) and probabilistic methods (multi-hypothesis).

\textbf{Deterministic methods} \cite{martinez2017simple,zhao2019semantic,zheng20213d,zhang2022mixste} predict a single definitive 3D pose for each input, which is straightforward and practical for real-world applications. 
Early works like SimpleBaseline \cite{martinez2017simple} employ fully-connected layers with residual connections to directly predict 3D joints from 2D detections.
Since the human skeleton can be naturally represented as a graph, various methods \cite {zhao2019semantic,cai2019exploiting,xu2021graph,zou2021modulated} based on Graph Convolutional Network (GCN) have been proposed.
Inspired by the success of Transformers \cite{Attention}, some work \cite{zheng20213d,strided,li2022mhformer,zhang2022mixste} focuses on model long-range dependencies and captures global correlations among joints. 

In contrast, \textbf{probabilistic methods} \cite{jahangiri2017generating,sharma2019monocular,li2019generating,oikarinen2021graphmdn} explicitly model the ambiguity of monocular 3D pose estimation by generating multiple plausible hypotheses. 
The early work of Jahangiri and Yuille~\cite{jahangiri2017generating} produces multiple 3D pose hypotheses using a hand-crafted anatomical generative model conditioned only on 2D joints.
Li and Lee \cite{li2019generating} introduced MDN \cite{bishop1994mixture} to generate pose hypotheses, while Oikarinen et al. \cite{oikarinen2021graphmdn} further enhanced this framework by integrating MDN with GCN to exploit the inherent graph structure of the human skeleton.
Sharma et al. \cite{sharma2019monocular} used a Conditional Variational AutoEncoder (CVAE) \cite{sohn2015learning} to obtain diverse 3D pose samples.
Wehrbein et al. \cite{wehrbein2021probabilistic} explored normalizing flows, and Li et al. \cite{li2020weakly} leveraged generative adversarial network (GAN) \cite{goodfellow2020generative} to model the distribution of 3D poses. 

Recently, diffusion-based models \cite{gong2023diffpose,shan2023diffusion,zhang2025champ} have been explored for 3D human pose estimation, leveraging their iterative denoising process to generate diverse and plausible 3D pose samples.
Representative works include DiffPose \cite{gong2023diffpose}, D3DP \cite{shan2023diffusion}, and CHAMP \cite{zhang2025champ}.
Given that multiple 3D poses can correspond to the same 2D observation, probabilistic modeling is particularly well suited for this task, and we adopt it in this work.

Although monocular 3D pose estimation has been extensively studied for humans \cite{liu2022recent,zheng2023deep}, its counterpart for animals remains far less explored. The limited availability of 3D annotations \cite{jiang2022animal} has driven most existing methods toward model-based formulations. These approaches fit the Skinned Multi-Animal Linear (SMAL) model \cite{zuffi20173d} to images to reconstruct the 3D shape of the animal \cite{zuffi2019three,biggs2020left,lyu2025animer,li2024learning,kaye2025dualpm}, from which the 3D pose is obtained by joint regression from the fitted mesh. The problem is further compounded by substantial morphological variation between species, making monocular 3D animal pose estimation considerably more challenging than its human counterpart.

\subsection{Flow Matching in Generative Modeling}
Diffusion models \cite{ho2020denoising,song2021denoising} have achieved significant success in tasks such as image \cite{dhariwal2021diffusion,ramesh2022hierarchical}, sound \cite{huang2023noise2music,liu2023audioldm}, and video generation \cite{fei2024dysen}.
They are typically formulated via SDEs and require multiple denoising steps for inference.
Recently, several works \cite{gong2023diffpose,shan2023diffusion,zhang2025champ} have leveraged diffusion models for 3D human pose estimation, where sampling from noise enables the generation of diverse and plausible pose hypotheses. 
However, diffusion-based inference is slow, as the inference stage requires many iterative denoising steps, making these methods impractical for real-time applications.

Recently, Flow Matching \cite{lipman2023flow,liu2023rectifiedflow} was used to model the velocity field that transports a simple base distribution to the data distribution via ODEs. FM has shown strong results on various other tasks \cite{wang2024semflow,wang2025editor,gui2025depthfm} while enabling faster sampling with much fewer steps. Its deterministic ODE trajectories allow for fewer sampling steps and yield lower latency, making it suitable for real-time applications.  In this work, we adopt conditional FM for 3D pose estimation, enabling efficient inference and multi-hypothesis generation via different noise initializations.

\begin{figure}[t]
    \centering
    \includegraphics[width=1.0\linewidth]{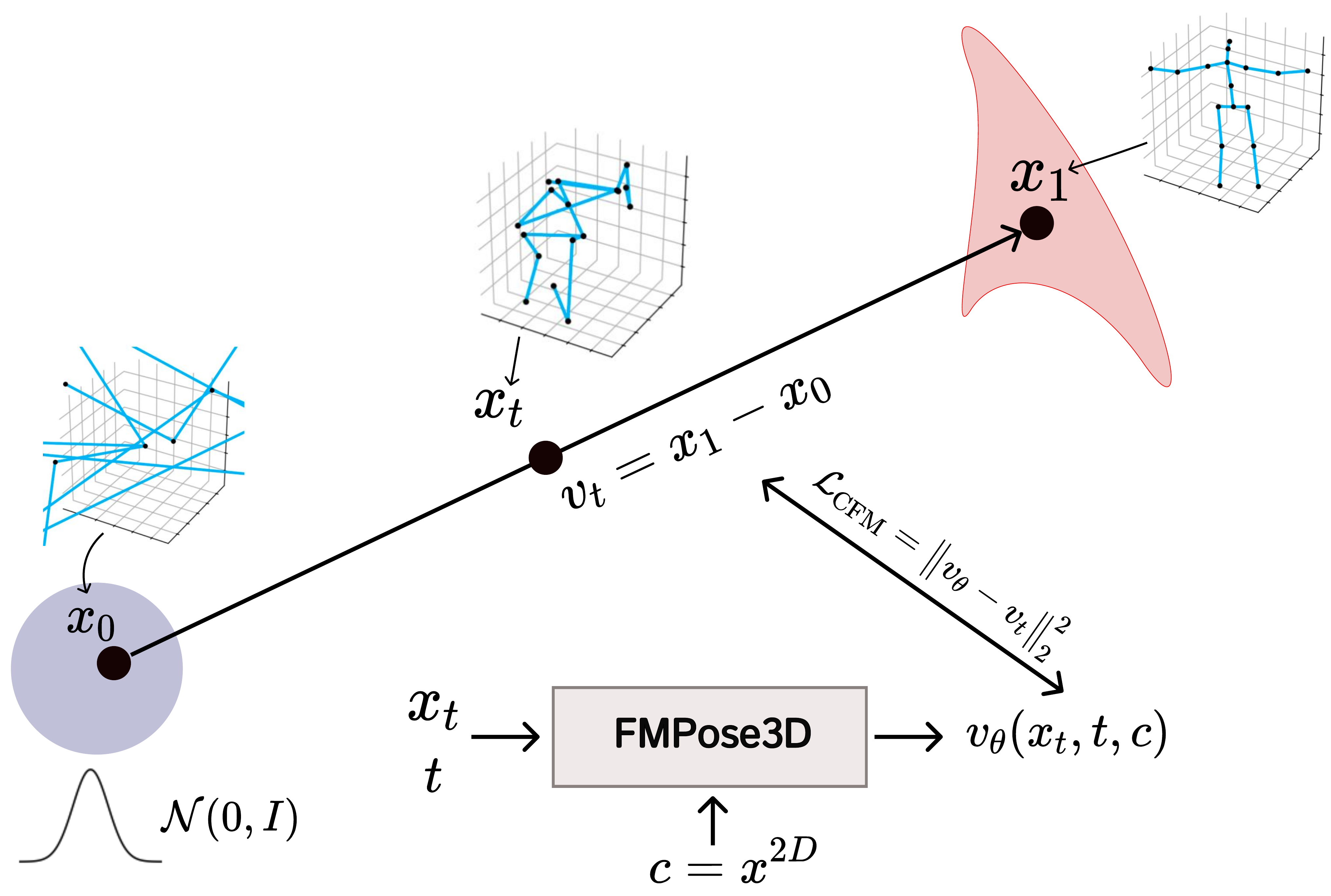}
    \caption{
    Overview of the training process.
    The process starts from a noise sample $x_0 \!\sim\! \mathcal{N}(0, I)$ and a ground-truth 3D pose $x_1$ from the training set. 
    An intermediate sample $x_t$ is obtained by linear interpolation between $x_0$ and $x_1$. 
The red region illustrates the valid 3D pose data manifold.
The network $v_\theta(x_t, t, c)$, conditioned on the 2D pose $c = x^{2D}$, is trained to predict the true velocity $v_t$. The Flow Matching loss $\mathcal{L}_{\text{CFM}} = \|v_\theta - v_t\|_2^2$ minimizes the discrepancy between the predicted and ground-truth velocities.}
    \label{fig:fmpose_training}
\end{figure}

\section{Methods}
\label{sec:method}

\subsection{3D Pose Estimation via Flow Matching}
In this work, we propose \textbf{FMPose3D}, a generative framework for 3D pose estimation built upon flow matching. Instead of directly regressing a 3D pose from a 2D input,  we cast 3D pose estimation as a conditional distribution transport problem, where the 2D pose serves as the conditioning signal. The model learns a conditional velocity field that transports samples from a simple Gaussian base distribution to the target conditional distribution of plausible 3D poses given a 2D observation.

\textbf{Problem setup.}
Let $x^{2D}\!\in\!\mathbb{R}^{J\times 2}$ and $x_1\!\in\!\mathbb{R}^{J\times 3}$ denote a paired 2D and 3D pose with $J$ joints, and let the 2D pose serve as the condition $c = x^{2D}$. 
Let $p_0=\mathcal{N}(0,I)$ be a standard Gaussian base distribution over $\mathbb{R}^{J\times 3}$, and sample $x_0 \sim p_0$. FMPose3D learns a conditional velocity field that deterministically transports samples from $p_0$ to the conditional data distribution $p_{\text{data}}(x_1\!\mid\!c)$.

\textbf{Linear interpolation path.}
Given a noise sample $x_0\!\sim\!p_0$ and a target 3D pose $x_1$, we define the linear interpolation:
\begin{equation}
x_t \;=\; (1-t)\,x_0 \;+\; t\,x_1, \quad t \in [0,1).
\label{eq:cond-path}
\end{equation}
During training, we sample $t \sim \mathcal{U}[0,1]$ and use the corresponding $x_t$ as an intermediate state, so that the velocity field $v_\theta(x_t, t, c)$ learns to move points along this path toward the target pose $x_1$.

\textbf{Target velocity.}
Differentiating Eq.~\eqref{eq:cond-path} with respect to $t$ yields the instantaneous velocity along the interpolation path:
\begin{equation}
v_t \;=\; \frac{d x_t}{d t} \;=\; x_1 - x_0 .
\end{equation}
This path-constant vector serves as the ground-truth target used to supervise the conditional velocity field $v_\theta(x_t, t, c)$ at each intermediate state $x_t$.

\textbf{Conditional flow matching loss.}
A neural network $v_\theta(x,t,c)$ is trained to approximate the target velocity $v_t$ defined above, where $c=x^{2D}$.
The Conditional Flow Matching (CFM) objective minimizes the expected squared error between the predicted and target velocities:
\begin{equation}
\mathcal{L}_{\text{CFM}}(\theta)
=
\mathbb{E}_{x_0\sim p_0,\, t\sim\mathcal{U}[0,1)}
\!\left[
\big\|
v_\theta(x_t, t, c)
- (x_1 - x_0)
\big\|_2^2
\right].
\label{eq:cfm-loss}
\end{equation}

\textbf{Overview of the training process.}
The training pipeline is illustrated in Figure~\ref{fig:fmpose_training}. At each iteration, a paired sample $(x^{2D},x_1)$ is drawn from the dataset, while Gaussian noise $x_0\!\sim\!\mathcal{N}(0,I)$ and a time $t\!\sim\!\mathcal{U}[0,1)$ are sampled. The intermediate state $x_t$ is then constructed by linear interpolation as in Eq.~\eqref{eq:cond-path}. Given the current state $(x_t,t,c)$, the network predicts the instantaneous velocity $v_\theta(x_t,t,c)$, and the target velocity is the path-constant vector $v_t=x_1-x_0$. The per-sample loss $\mathcal{L}_{\text{CFM}}=\|v_\theta-v_t\|_2^2$ penalizes the discrepancy between the predicted and target velocities. Intuitively, training learns a conditional velocity field that transports samples from Gaussian noise toward the 3D pose distribution under the 2D condition.

\begin{figure*}[!t]
\centering
\centerline{\includegraphics[width=1.0\linewidth]{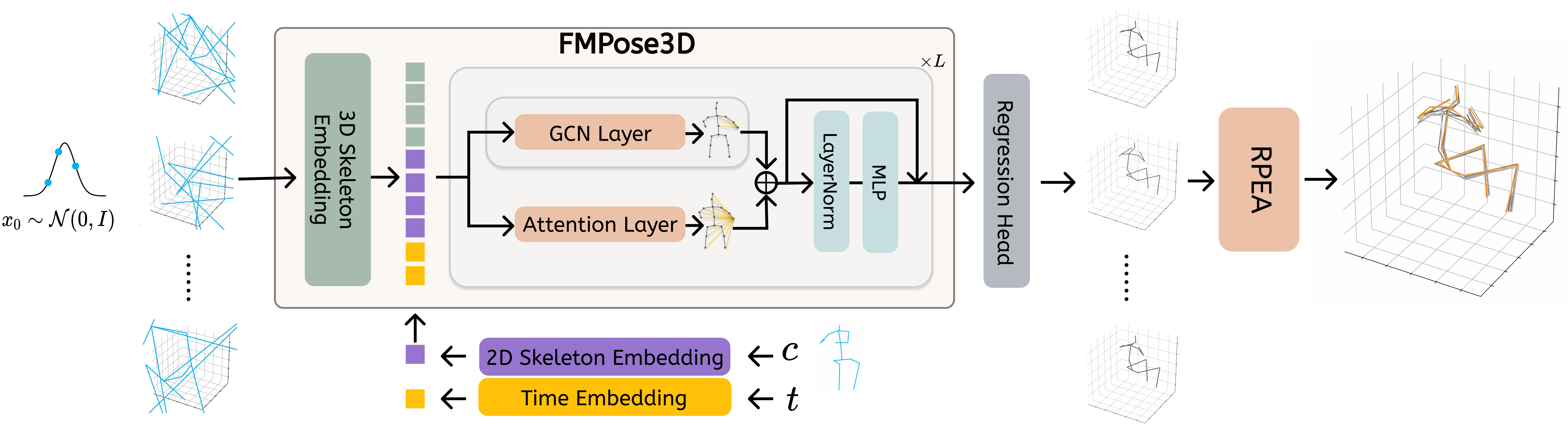}}
\caption{Illustration of multi-hypothesis generation and aggregation during inference.}
\label{fig:inference}
\end{figure*}

\textbf{Sampling (inference).}
After training, the learned conditional velocity field $v_\theta(x,t,c)$ defines the ODE:
\begin{equation}
\frac{d x_t}{d t} \;=\; v_\theta(x_t, t, c), \qquad x_0 \sim p_0 = \mathcal{N}(0,I),
\end{equation}
whose deterministic flow transports samples from the source distribution $p_0$ to the conditional distribution $p_\theta(x \mid c)$.

At inference time, given a 2D pose $c = x^{2D}$, we predict a 3D pose $\hat{x}^{3D}$ by solving this ODE. Sampling an initial noise $x_0 \sim \mathcal{N}(0,I)$ and integrating the velocity field from $t=0$ to $1$, we obtain the following integral expression:
\begin{equation}
\hat{x}^{3D} \;=\; x_0 \;+\; \int_{0}^{1} v_\theta\!\big(x_t, t, c\big)\, dt.
\label{eq:integral_form}
\end{equation}

In practice, the integral is approximated with $S$ discrete steps (step size $h=1/S$). Using explicit Euler, the update is
\begin{equation}
x_{\,t+\frac{1}{S}} \;=\; x_t \;+\; \frac{1}{S}\, v_\theta(x_t, t, c),
\qquad t \in \Big\{0, \tfrac{1}{S}, \dots, 1-\tfrac{1}{S}\Big\}.
\label{eq:euler_forward}
\end{equation}
This process transports the noisy initialization $x_0$ toward a plausible 3D pose $\hat{x}^{3D}$ that is consistent with the given 2D observation.

Algorithm~\ref{alg:fmpose} summarizes the complete training and single-hypothesis inference procedure of our FMPose3D.

\subsection{Reprojection-based Posterior Expectation Aggregation (RPEA)}

As shown in Figure~\ref{fig:inference}, at inference time, given a 2D pose $X^{2D}$, we can draw $N$ noise samples to generate $N$ 3D pose hypotheses. To obtain a single accurate prediction, we introduce a new aggregation framework, \textbf{Reprojection-based Posterior Expectation Aggregation (RPEA)}. Our method is motivated by Bayesian decision theory \cite{berger2013statistical} and aims to provide a more robust solution to the multi-hypothesis aggregation problem.

During training, the model implicitly approximates the uncertainty over the 3D pose, which is described by the posterior distribution $p(X^{3D} | X^{2D})$. Our objective is to derive a single, unique 3D pose estimate $\hat{X}^{3D}$ from this distribution. According to Bayesian decision theory, a rational objective is to select an estimate $\hat{X}^{3D}$ that minimizes the Expected Loss.
With Mean Squared Error (MSE) as the decision loss function, the corresponding risk is defined as:
\begin{equation}
    R(\hat{X}^{3D}) = \mathbb{E}\left[ ||\hat{X}^{3D} - X^{3D}||^2 \right].
\end{equation}

\begin{algorithm}[t]
\caption{Training and inference of FMPose3D}
\label{alg:fmpose}
\KwIn{
  Training set $\mathcal{D} = \{(x^{2D}, x^{3D})\}$ with
  $x^{2D}\!\in\!\mathbb{R}^{J\times 2}$ and $x^{3D}\!\in\!\mathbb{R}^{J\times 3}$; 
  number of training iterations $T_{\mathrm{train}}$; number of Euler steps $S$ used at inference.
}
\KwOut{
 Trained FMPose3D parameters $\theta$ (velocity field $v_\theta$); at inference time, the predicted 3D pose $\hat{x}^{3D}$ for a given 2D pose.
}
\textbf{Training phase:}\\[2pt]
\For{$iter = 1$ \KwTo $T_{\mathrm{train}}$}{
    Sample a mini-batch of data pairs $(x^{2D}, x^{3D})$ from $\mathcal{D}$\;
    Set $x_1 \leftarrow x^{3D}$ and condition $c \leftarrow x^{2D}$\;
    Sample $x_0 \!\sim\! \mathcal{N}(0,I)$ and $t \!\sim\! \mathcal{U}[0,1)$\;
    Compute interpolated states $x_t \leftarrow (1-t)\,x_0 + t\,x_1$\;
    Predict velocities $v^{\mathrm{pred}} \leftarrow v_\theta(x_t, t, c)$\;
    Set target velocities $v^{\mathrm{target}} \leftarrow x_1 - x_0$\;
    Compute the CFM loss on the batch
    $\mathcal{L}_{\mathrm{CFM}} \leftarrow \|v^{\mathrm{pred}} - v^{\mathrm{target}}\|_2^2$\;
    Update parameters $\theta \leftarrow \mathrm{Adam}(\theta, \nabla_\theta \mathcal{L}_{\mathrm{CFM}})$\;
}

\textbf{Inference phase (given trained $\theta$):}\\[2pt]
Given a 2D pose $x^{2D}$ and the number of Euler steps $S$ \;
Set condition $c \leftarrow x^{2D}$ and sample $x_0 \sim \mathcal{N}(0,I)$\;
Initialize $x \leftarrow x_0$\;
\For{$k = 0$ \KwTo $S-1$}{
    $t \leftarrow k/S$\;
    $x \leftarrow x + \frac{1}{S}\, v_\theta(x, t, c)$\;
}
\Return $\hat{x}^{3D} \leftarrow x$\;
\end{algorithm}

Our goal is to find the estimator that minimizes this risk, known as the Minimum Mean Squared Error (MMSE) estimator:
\begin{equation}
    \hat{X}^{MMSE} = \arg\min_{\hat{X}^{3D}} R(\hat{X}^{3D}).
\end{equation}

According to Bayesian decision theory, the MMSE estimator is precisely the expectation of the posterior distribution:
\begin{equation}
    \hat{X}^{MMSE} = E[X^{3D}|X^{2D}] = \int_{\mathbb{R}^d} X^{3D} \cdot p(X^{3D}|X^{2D}) \, dX^{3D}.
\end{equation}

However, the analytical form of the posterior distribution $p(X^{3D} | X^{2D})$ is unknown, making it impossible to compute the above integral directly. 
To approximate this distribution, we make a reasonable assumption that the posterior probability of a 3D hypothesis is proportional to the exponential of its negative 2D re-projection loss $L$:
\begin{equation}
    p(H_i|X^{2D}) \propto \exp(-\alpha \cdot L(H_i, X^{2D})),
\end{equation}
where $\alpha$ is a fixed temperature hyperparameter. Under this assumption, we can approximate the posterior expectation using a weighted Monte Carlo estimator, where the weights are determined by the above pseudo-likelihood.

Here we show how the joint-wise RPEA performs. First, we use the pre-trained model to draw $N$ samples from the learned conditional distribution $p_\theta(X^{3D}\mid X^{2D})$, yielding $N$ 3D pose hypotheses $\{H_1, H_2, ..., H_N\}$.
Our RPEA method applies the principle of posterior expectation estimation independently to each joint. Specifically, the final position of each joint is robustly estimated via the following two steps:
\begin{enumerate}
    \item \textbf{Filtering:} For the $j$-th joint, we consider its $N$ candidate positions $\{H_{1,j}, ..., H_{N,j}\}$ from the $N$ hypotheses. Let $L_{i,j}$ denote the re-projection loss associated with the $j$-th joint of hypothesis $H_i$. We rank these candidates by their losses $\{L_{1,j}, ..., L_{N,j}\}$ and select the Top-K candidates with the lowest losses to form a high-likelihood candidate set $\mathcal{H}_{K,j}$.
    
    \item \textbf{Weighted Aggregation:} We then compute the weighted average over this high-quality candidate set to approximate the posterior expectation of the joint's position.
\end{enumerate}

The final estimated position of the $j$-th joint, $\hat{X}^{RPEA}_{j}$, is given by:
\begin{equation}
\label{eq:rpea_final_en}
\begin{aligned}
\hat{X}^{RPEA}_{j} &= \sum_{H_{i,j} \in \mathcal{H}_{K,j}} w_{i,j} \cdot H_{i,j}, \\
\text{where}\quad 
w_{i,j} &= \frac{\exp(-\alpha L_{i,j})}{
\sum_{H_{k,j} \in \mathcal{H}_{K,j}}\exp(-\alpha L_{k,j})}.
\end{aligned}
\end{equation}

The final 3D pose $\hat{X}^{RPEA}$ is then constructed by assembling the estimated positions of all joints $\{\hat{X}^{RPEA}_{1}, ..., \hat{X}^{RPEA}_{J}\}$.

\subsection{Architecture}

FMPose3D parameterizes the conditional velocity field $v_\theta(x_t, t, c)$ used in the ODE of Sec.~\ref{sec:method}, where $x_t\!\in\!\mathbb{R}^{J\times 3}$ is the current 3D pose state (with $x_0$ initialized from Gaussian noise at $t=0$), $c = x^{2D}\!\in\!\mathbb{R}^{J\times 2}$ is the 2D pose condition, and $t\!\in\![0,1]$ is the integration time. At each ODE step, FMPose3D takes $(x_t, x^{2D}, t)$ as input and predicts the instantaneous velocity $v_\theta(x_t, t, c)$.

As shown in Figure~\ref{fig:inference}, we first apply three embedding layers: a 3D skeleton embedding layer that maps the current 3D pose state $x_t$ to per-joint features, a 2D skeleton embedding layer for the 2D pose $x^{2D}$, and a time embedding layer for the time $t$. These embeddings transform the inputs into latent feature space. All three embedding layers are implemented as lightweight MLPs. The three embeddings are then concatenated to form a joint-wise feature tensor $F_t \in \mathbb{R}^{J\times D}$, which serves as the input to the backbone.
The backbone of FMPose3D mainly consists of two parallel branches: a local GCN branch and a global self-attention branch, designed to capture complementary spatial relationships among human joints. The local branch, implemented with the GCN \cite{kipf2017semisupervised} layer in the upper path, treats the human skeleton as a graph with joints as nodes, effectively encoding topological information and modeling short-range dependencies between neighboring joints. The global branch, implemented with the attention~\cite{Attention} layer in the lower path, captures long-range contextual interactions between non-adjacent joints, providing a holistic structural understanding of the 3D pose.
The outputs of the two branches are concatenated and passed through a LayerNorm \cite{ba2016layer} and an MLP. Finally, a regression head (an MLP applied per joint) maps the backbone features to the predicted velocity $v_\theta(x_t, t, c) \in \mathbb{R}^{J\times 3}$.
% For details about GCN and attention, please see Suppl. Sec. \ref{sec:supp_attn_gcn}.

\section{Experiments}

\begin{table*}[t]
\caption{Quantitative comparison with the state-of-the-art methods on Human3.6M under MPJPE. The detected 2D pose is used as input. $N$ denotes the number of hypotheses.
\textcolor{red}{Red}: Best. \textcolor{blue}{Blue}: Second Best.
\colorbox{grey!10}{Grey}: our method.
}
% The top two best results for each action are highlighted in bold and underlined, respectively. 
\resizebox{\textwidth}{!}{
\begin{tabular}{lr|ccccccccccccccc|c}
\toprule
% \multicolumn{17}{c}{\textbf{Deterministic Method}} \\
% \cmidrule(lr){1-17}
\multicolumn{2}{c|}{\textbf{Deterministic Method}} & Dire. & Disc. & Eat & Greet & Phone & Photo & Pose & Purch. & Sit & SitD. & Smoke & Wait & WalkD. & Walk & WalkT. & Avg $\downarrow$  \\
\midrule
SimpleBaseline \cite{martinez2017simple} & ICCV'17 &51.8 & 56.2& 58.1 & 59.0 & 69.5 & 78.4 & 55.2& 58.1& 74.0 & 94.6 & 62.3 & 59.1 & 65.1& 49.5 & 52.4 & 62.9\\
VideoPose3D \cite{pavllo20193d} & CVPR'19 & 47.1& 50.6 & 49.0 & 51.8 &53.6&61.4 &49.4 &47.4& 59.3 &67.4& 52.4& 49.5& 55.3 &39.5 &42.7 &51.8\\
LCN \cite{ci2019optimizing} & ICCV'21 & 46.8 & 52.3 & 44.7 & 50.4 & 52.9 & 68.9 & 49.6 & 46.4 & 60.2 & 78.9 & 51.2 & 50.0 & 54.8 & 40.4 & 43.3 & 52.7\\
SRNet \cite{zeng2020srnet} & ECCV'20 & 44.5 & 48.2 & 47.1 & 47.8 & 51.2 & \textcolor{blue}{56.8} & 50.1 & 45.6 & 59.9 & 66.4 & 52.1 & 45.3 & 54.2 & 39.1 & 40.3 & 49.9 \\
GraphSH \cite{xu2021graph}  & CVPR'21 & 45.2 & 49.9 & 47.5 & 50.9 & 54.9 & 66.1 & 48.5 & 46.3 & 59.7 & 71.5 & 51.4 & 48.6 & 53.9 & 39.9 & 44.1 & 51.9 \\
% % % MGCN \cite{zou2021modulated} ICCV'21 $\S$ & 45.4 & 49.2 &  45.7 & 49.4 & 50.4 & 58.2 & 47.9 & 46.0 & 57.5 & 63.0 & 49.7 & 46.6 & 52.2 & 38.9 & 40.8 & 49.4 \\
GraFormer \cite{zhao2022graformer} & CVPR'22 & 45.2 &50.8 & 48.0 & 50.0 & 54.9& 65.0 & 48.2& 47.1 & 60.2 & 70.0&51.6 & 48.7 & 54.1 & 39.7 & 43.1 & 51.8 \\
UGRN \cite{li2023pose} & AAAI'23 & 47.9 & 50.0 & 47.1 & 51.3 & 51.2 & 59.5 & 48.7 & 46.9 & \textcolor{blue}{56.0} & 61.9 & 51.1 & 48.9 & 54.3 & 40.0 & 42.9 & 50.5 \\
MLP-JCG \cite{tang2023mlp} & TMM'23 & 43.8 & \textcolor{red}{46.7} & 46.9 & 48.9 & \textcolor{blue}{50.3} & 60.1 & \textcolor{blue}{45.7} & \textcolor{blue}{43.9} & \textcolor{blue}{56.0} & 73.7 & \textcolor{blue}{48.9} & 48.2 & \textcolor{blue}{50.9} & 39.9 &41.5 &49.7 \\
PerturbPE \cite{azizi2024occlusion} & ECCV'24  & - & - & - & - & - & - & - & - & - & - & - & - & - & - & - & 50.8 \\
\toprule
% % \multicolumn{17}{c}{\textbf{Probabilistic Method}} \\
% % \cmidrule(lr){1-17}
\multicolumn{2}{c|}{\textbf{Probabilistic Method}} & Dire. & Disc. & Eat & Greet & Phone & Photo & Pose & Purch. & Sit & SitD. & Smoke & Wait & WalkD. & Walk & WalkT. & Avg $\downarrow$  \\
\midrule
CVAE ($N$=200) \cite{sharma2019monocular} & ICCV'19 & 48.6 & 54.5 & 54.2 & 55.7 & 62.6 & 72.0 & 50.5 & 54.3 & 70.0 & 78.3 & 58.1 & 55.4 & 61.4 & 45.2 & 49.7 & 58.0 \\
GAN ($N$=10) \cite{li2020weakly} & BMVC'20 & 66.0 & 74.7 & 71.1 & 80.6 & 81.1 & 93.0 & 73.2 & 83.7 & 90.0 & 117.4 & 75.8 & 79.3 & 82.1 & 74.4 & 77.8 & 80.9 \\
GraphMDN ($N$=5) \cite{oikarinen2021graphmdn} & IJCNN'21 & 51.9 & 56.1 & 55.3 & 58.0 & 63.5 & 75.1 & 53.3 & 56.5 & 69.4 & 92.7 & 60.1 & 58.0 & 65.5 & 49.8 & 53.6 & 61.3 \\
NF ($N$=1) \cite{wehrbein2021probabilistic} & ICCV'21 & 52.4 & 60.2 & 57.8 & 57.4 & 65.7 & 74.1 & 56.2 & 59.1 & 69.3 & 78.0 & 61.2 & 63.7 & 67.0 & 50.0 & 54.9 & 61.8 \\
DiffPose ($N$=5) \cite{gong2023diffpose} & CVPR'23  & \textcolor{red}{42.8} & 49.1 & \textcolor{blue}{45.2} & \textcolor{blue}{48.7} & 52.1 & 63.5 & 46.3 & 45.2 & 58.6 & 66.3 & 50.4 & 47.6 & 52.0 & \textcolor{blue}{37.6} & \textcolor{blue}{40.2} & 49.7 \\
ProPose ($N$=1) \cite{han2025propose} & AAAI'25 & - & - & - & - & - & - & - & - & - & - & - & - & - & - & - & 51.9 \\
% \midrule
% FMPose3D (Ours) & 46.2 & 49.7 & 46.3 & 49.8 & 51.3 & 57.8 & 47.7 & 45.5 & 58.5 & 61.8 & 50.2 & 47.1 & 52.7 & 39.3 & 42.1 & 49.7 \\
\rowcolor{grey!10} FMPose3D ($N$=2) (Ours)& & 45.7 & 49.2 & 46.0 & 49.4 & 50.9 & 57.3 & 47.2 & 45.0 & 57.9 & \textcolor{blue}{61.5} & 49.8 & \textcolor{blue}{46.8} & 52.1 & 38.9 & 41.6 & \textcolor{blue}{49.3} \\
\rowcolor{grey!10} FMPose3D ($N$=40) (Ours) & & \textcolor{blue}{43.5} & \textcolor{blue}{47.2} & \textcolor{red}{44.4} & \textcolor{red}{47.7} & \textcolor{red}{48.9} & \textcolor{red}{55.1} & \textcolor{red}{45.5} & \textcolor{red}{42.7} & \textcolor{red}{55.7} & \textcolor{red}{59.4} & \textcolor{red}{47.9} & \textcolor{red}{45.1} & \textcolor{red}{49.8} & \textcolor{red}{37.1} & \textcolor{red}{39.6} &  \textcolor{red}{47.3} \\
\bottomrule
\end{tabular}
}
\label{tab:human3.6M_cpn}
\end{table*}

\subsection{Datasets and Evaluation Metrics}
Here, we provide a more detailed description of the datasets and evaluation metrics.

\textbf{Human3.6M} \cite{h36m} is the most representative benchmark for the estimation of 3D human poses.
It contains 3.6 million video frames captured from four synchronized cameras at 50Hz in an indoor environment.
There are 11 professional actors performing 15 actions, such as greeting, phoning, and sitting.
Following previous works \cite{pavllo20193d, cai2019exploiting}, we train our model on five subjects (S1, S5, S6, S7, S8) and test it on two subjects (S9 and S11). 
The performance is evaluated by two common metrics: MPJPE (Mean Per-Joint Position Error) is the mean Euclidean distance between the predicted joints and the ground truth in millimeters. 
P-MPJPE is the MPJPE after rigid alignment with the ground truth in translation, rotation, and scale.

\textbf{MPI-INF-3DHP} \cite{mehta2017monocular} is a more challenging 3D pose dataset that contains both complex indoor and outdoor scenes.
There are 8 actors performing 8 actions from 14 camera views, which cover a greater diversity of poses. 
Its test set consists of three different scenes: studio with green screen (GS), studio without green screen (noGS), and outdoor scene (Outdoor). Following \cite{zou2021modulated,zeng2021learning,pavllo20193d}, we use Percentage of Correct Keypoints (PCK) with a threshold of 150mm and the Area Under Curve (AUC) for a range of PCK thresholds for evaluation.

 \textbf{Animal3D} \cite{xu2023animal3d} is built on a set of images adapted from ImageNet \cite{deng2009imagenet} and COCO \cite{lin2014microsoftcoco} datasets. Annotators manually labeled 2D keypoints and silhouettes to aid an optimization-based fitting process, producing pseudo SMAL \cite{zuffi20173d} labels. The dataset includes 3.4k instances with 40 species. In this study, we used 3D keypoints derived from SMAL models along with reprojected 2D keypoints for training and evaluation purposes.

 \textbf{CtrlAni3D} \cite{lyu2025animer} is a synthetic 3D animal dataset generated using ControlNet \cite{zhang2023adding} by conditioning textual descriptions of animal behaviors on SMAL structures and rendering them into photorealistic images. The dataset comprises 9.7k images spanning 10 species, with each image annotated with pixel-aligned SMAL meshes, enabling precise 3D shape and pose analysis.

\begin{table}[!t]
    \centering
     \caption{Quantitative comparisons with state-of-the-art methods on MPI-INF-3DHP.}
    \setlength\tabcolsep{1.80mm}
    \footnotesize
    \resizebox{1.0 \linewidth}{!}{
    \begin{tabular}{l|c|c|c|c|c}
    \toprule 
    Method  & GS $\uparrow$ & noGS $\uparrow$&  Outdoor $\uparrow$ & All PCK $\uparrow$& All AUC $\uparrow$\\
    \midrule
    SimpleBaseline \cite{martinez2017simple} & 49.8 & 42.5 & 31.2 &42.5 & 17.0 \\
    GraphSH \cite{xu2021graph} & 81.5 & 81.7 & 75.2 & 80.1 & 45.8 \\ 
    Zeng \textit{et al.} ~\cite{zeng2021learning} & - &- & 84.6 & 82.1 & 46.2 \\
    GraFormer \cite{zhao2022graformer} & 80.1 &77.9 &74.1 &79.0& 43.8 \\
    UGRN \cite{li2023pose} & \textcolor{red}{86.2} & 84.7 & 81.9 & 84.1 & \textcolor{blue}{53.7} \\
    PerturbPE \cite{azizi2024occlusion} & 80.0 & 79.0 & 84.0 & 82.0 & - \\
    ProPose \cite{han2025propose} & 83.9 & 85.5 & 83.4 & 84.4 & 52.1 \\
    \midrule
\rowcolor{grey!10}    FMPose3D ($N$=2) (Ours) & 85.7 & \textcolor{blue}{86.4} & \textcolor{blue}{85.6} & \textcolor{blue}{85.9} & \textcolor{blue}{53.7} \\
\rowcolor{grey!10}    FMPose3D ($N$=20) (Ours) & \textcolor{blue}{86.1} & \textcolor{red}{87.1} & \textcolor{red}{86.5} & \textcolor{red}{86.4} & \textcolor{red}{54.6} \\
    \bottomrule
    \end{tabular}
    }
    \label{tab:3dhp}
\end{table}

\subsection{Implementation Details}
% flip Augmentation
% \noindent \textbf{Humans}. 
During evaluation, we employ horizontal flip augmentation, following prior works \cite{cai2019exploiting, pavllo20193d, zeng2021learning}. Instead of simply averaging the predictions from the original and flipped inputs, 
we treat the two as separate hypotheses and feed them into our RPEA module to obtain the final 3D pose.
% we integrate this augmentation strategy directly into our multi-hypothesis framework: the prediction from the original 2D pose and the prediction from its flipped counterpart are treated as two distinct hypotheses, which are then fed into our proposed RPEA module to compute the final aggregated 3D pose. 
We refer to this strategy as \textbf{Flipped-Hypothesis Aggregation (FHA).} 
Following \cite{xu2021graph,zeng2021learning,cai2019exploiting,pavllo20193d}, we use 2D pose detected by CPN \cite{chen2018cascaded} for Human3.6M,
and ground truth 2D pose for MPI-INF-3DHP.
During inference, we set the number of ODE steps to $S=3$.
More details are provided in Suppl. Sec \ref{sec:suppl_implementation}.

\begin{table}[!t]
    \centering
     \caption{Quantitative comparisons with state-of-the-art methods on Animal3D and CtrlAni3D.}
    \setlength\tabcolsep{5.60mm}
    \footnotesize
\begin{tabular}{c|cc}
\toprule
Dataset & Animal3D & CtrlAni3D \\ \cline{1-3} 
Metric  & P-MPJPE $\downarrow$   & P-MPJPE $\downarrow$   \\ \hline
HMR \cite{kanazawa2018end}     & 123.5    & 123.5     \\
WLDO \cite{biggs2020left}   & 112.3    & 71.5      \\
HMR2.0 \cite{goel2023humans}  & 94.1     & 60.9      \\
AniMer \cite{lyu2025animer}  & \textcolor{blue}{80.4}     & \textcolor{blue}{44.1}      \\ 
\hline
\rowcolor{grey!10} Ours    & \textcolor{red}{61.5}     & \textcolor{red}{44.0}          \\ 
\bottomrule
\end{tabular}
\label{tab:animal}
\end{table}

\subsection{Comparison with State-of-the-Art Methods}

\textbf{Human3.6M.}
Table~\ref{tab:human3.6M_cpn} presents the MPJPE comparison on Human3.6M between our FMPose3D and prior state-of-the-art deterministic (Table \ref{tab:human3.6M_cpn}, top) and probabilistic (Table \ref{tab:human3.6M_cpn}, bottom) methods.
% Following \cite{cai2019exploiting, xu2021graph}, the 2D pose detected by CPN \cite{chen2018cascaded} is used as input for training and testing.
Our baseline FMPose3D already achieves competitive accuracy at 49.3~mm MPJPE. 
Leveraging its generative capability, we apply \textbf{FHA} and sample 40 hypotheses per pose, with 20 drawn from the original 2D input and 20 from its horizontally flipped counterpart. 
% A key advantage of FMPose3D lies in its generative capability, which enables sampling multiple plausible 3D poses conditioned on the same 2D input.
% To leverage this capability, we apply \textbf{FHA} and generate 40 hypotheses for each pose, where 20 are sampled from the original 2D input and the remaining 20 are sampled from its horizontally flipped counterpart.
These hypotheses are then aggregated using the proposed \textbf{RPEA} module (joint-wise), which further improves performance: \textbf{MPJPE decreases from 49.3~mm to 47.3~mm}, surpassing DiffPose~\cite{gong2023diffpose} (49.7~mm) by a notable margin (approximately 4.8\% relative improvement).
For P-MPJPE, we also achieve the best overall performance (see Suppl. Sec. \ref{sec:supp_h36m_p2}).

\textbf{MPI-INF-3DHP.}
To evaluate the generalization capability of the proposed FMPose3D, we further compare our approach with previous state-of-the-art methods in a cross-dataset setting. Specifically, we train our model solely on the Human3.6M dataset and directly test it on MPI-INF-3DHP without any fine-tuning.
As shown in Table \ref{tab:3dhp},
FMPose3D achieves the best overall performance in both PCK and AUC, consistently surpassing existing methods.
These results demonstrate the strong generalization ability of our approach to unseen scenarios.
See Suppl. Sec. \ref{sec:supp_3dpw} for results on another challenging in-the-wild dataset, 3DPW~\cite{pw3d}.

\textbf{Animal3D and CtrlAni3D.} 
% We compare our 3D estimation performance with state-of-the-art shape estimation methods that recover 3D animal shapes from image inputs, where 3D keypoints are derived from the predicted SMAL models. Due to the lack of single-view 3D pose lifting methods for animals, these shape-based approaches serve as our baselines. Our model is trained jointly on the two datasets and evaluated on each dataset separately. As shown in Table \ref{tab:animal}, our method achieves the best overall performance across both Animal3D and CtrlAni3D, with the lowest error on Animal3D and competitive accuracy on CtrlAni3D, demonstrating that our method generalizes well across datasets with different visual domains and species distributions. It is worth noting that the reported performance is obtained solely from our FMPose3D without incorporating the RPEA component.
% The consistent performance across datasets highlights the generality of our approach
We further evaluate our method on two animal pose datasets. 
Our model is trained jointly on Animal3D and CtrlAni3D and evaluated on each dataset separately. 
As shown in Table~\ref{tab:animal}, it achieves the best overall accuracy on both datasets.
Despite not using the RPEA module here, our model still surpasses shape-based baselines, demonstrating the robustness and adaptability of our framework.

\subsection{Ablation Study}

\begin{table}  
    \footnotesize
    \centering
    \caption{
    Ablation study on different model designs. \emph{Serial}: GCN followed by Attention (GCN$\rightarrow$Attn). 
    \emph{Parallel}: GCN and Attention are computed in two branches and fused.
    }
    \setlength\tabcolsep{2.6mm}
    \footnotesize
    \begin{tabular}{ccc|c}    
    \toprule
     Attention  & GCN & Connection & MPJPE $\downarrow$ \\
    \midrule
    \cmark &   &   -   & 50.9  \\ 
      &  \cmark  &    -  & 50.1  \\ 
    \cmark      &     \cmark  &  Serial  &  50.5  \\
        \cmark      &     \cmark  & Parallel  &  \textbf{49.3}  \\
    \bottomrule
    \end{tabular}    
    \label{tab:ab_model_design}
\end{table}

\begin{figure}[t]
    \centering
    \includegraphics[width=0.95\linewidth]{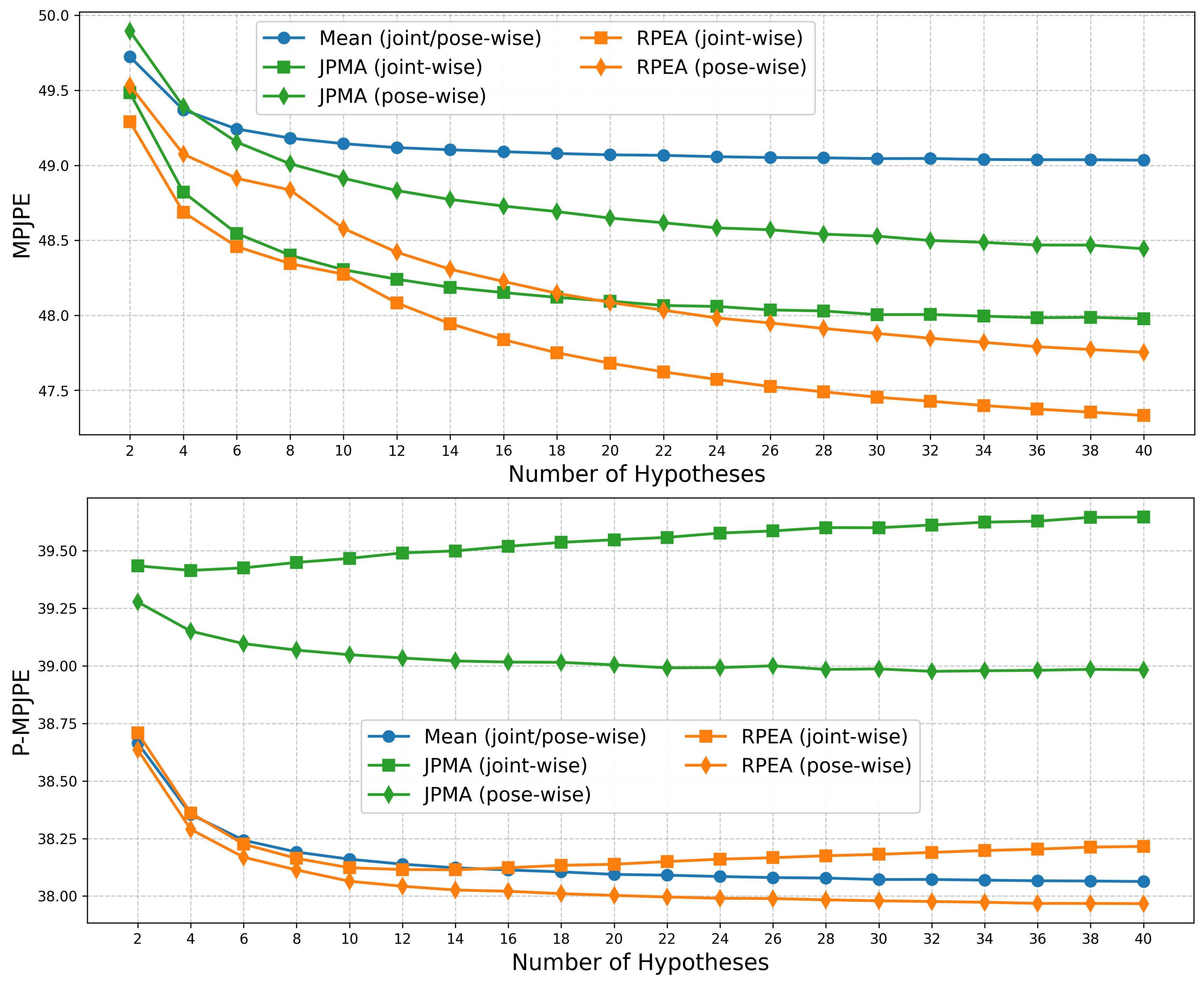}
    \caption{
    Comparison of different aggregation strategies on the Human3.6M test set. 
The top plot reports MPJPE, while the bottom plot shows P-MPJPE.
    }
    \label{fig:aggregation}
\end{figure}

\textbf{Impact of Model Design.}
We ablate backbone variants on Human3.6M using detected 2D poses (Table~\ref{tab:ab_model_design}).
A serial composition of the local GCN branch followed by the global self-attention branch (GCN$\rightarrow$Attention) yields limited improvement, as the sequential structure limits the ability to explore complementarity between local and global cues. In contrast, the parallel design that processes the two branches separately and fuses them at the feature level achieves the best performance, demonstrating the effectiveness of jointly leveraging local and global representations.

\begin{figure*}[!t]
\centering
\centerline{\includegraphics[width=1.0\linewidth]{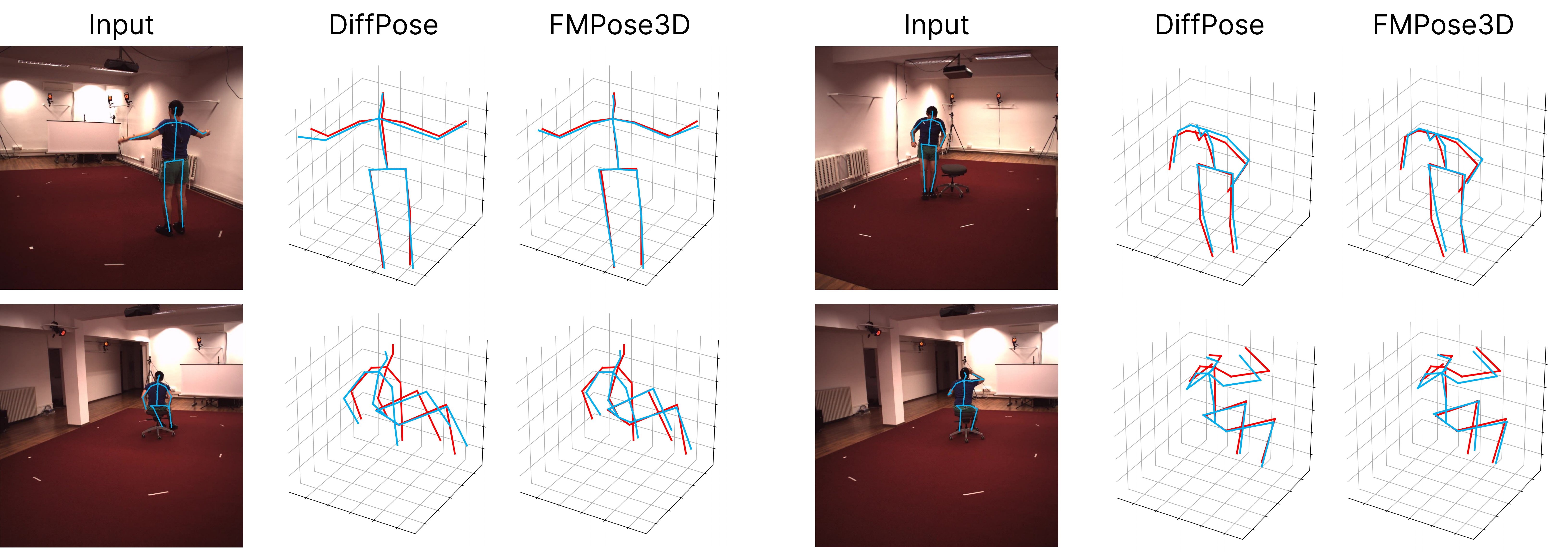}}
\caption
{Qualitative comparison of DiffPose \cite{gong2023diffpose} and FMPose3D on Human3.6M.
The blue pose represents the predicted results, while the red pose represents the ground truth.
}
\label{fig:qualitative_vis_h36m}
\end{figure*}

\begin{table}[!t]
    \centering
    \caption{Inference speed. Frames per second (FPS) were measured on a single GeForce RTX 4090 GPU. $N$ denotes the number of hypothesis. For DiffPose~\cite{gong2023diffpose}, we follow the setting in the original paper with 50 reverse diffusion steps at inference.}
    \setlength\tabcolsep{3.5mm}
    \footnotesize
    \begin{tabular}{l|c|c|c}
    \toprule
    Method & Steps & $N$ & FPS \\
    \midrule
    DiffPose \cite{gong2023diffpose} (w/o DDIM) & 50 & 5 &  3.36 \\
    DiffPose \cite{gong2023diffpose} (DDIM) & 5 & 5 & 27.15 \\
    \midrule
    FMPose3D (Ours) & 3 &  1 & 160.11 \\
    FMPose3D (Ours) & 3 & 40 & 145.59 \\
\bottomrule
\end{tabular}
\label{tab:inference_speed}
\end{table}

\textbf{Comparison with Different Aggregation Strategies.}
We compare the proposed RPEA with two alternative aggregation strategies:
(i) Mean, which computes the average of all hypotheses, as adopted in DiffPose~\cite{gong2023diffpose}; and
(ii) JPMA \cite{shan2023diffusion}, which selects either (a) the best 3D point for each joint independently (joint-wise) or (b) the best entire pose hypothesis (pose-wise), based on the minimum 2D reprojection loss in each case.
We apply \textbf{FHA} and evaluate on Human3.6M with different numbers $N$ of generated hypotheses. Due to FHA, $N$ is always even.
The comparison results for MPJPE are shown in Figure \ref{fig:aggregation} (top).
We observe that the Mean strategy yields only marginal improvements as the number of hypotheses increases, indicating that simple averaging cannot effectively exploit the diversity among samples.
Joint-wise JPMA yields improvements at small $N$, however, its performance quickly saturates beyond $N = 12$, reflecting its limited capacity to further exploit additional hypotheses.
In contrast, our RPEA consistently achieves lower MPJPE and demonstrates a much stronger ability to utilize additional hypotheses for performance gains.
To further assess pose structure up to Procrustes alignment, we report P-MPJPE results in Figure~\ref{fig:aggregation} (bottom).
We find that joint-wise JPMA fails to improve P-MPJPE, as assembling joints from different pose hypotheses may disrupt body structure and lead to anatomically inconsistent poses.
In contrast, our joint-wise RPEA maintains structural coherence and achieves P-MPJPE comparable to Mean, confirming that it generates anatomically consistent and plausible 3D poses. Moreover, our pose-wise RPEA achieves the best P-MPJPE with slightly worse MPJPE.

The impact of integration steps and intermediate-state visualizations are provided in Suppl. Sec. \ref{sec:suppl_vis_diff_S} and Sec. \ref{sec:suppl_impact_steps}.

% \begin{figure}[!t]
% \centering
% \centerline{\includegraphics[width=1.0\linewidth]{figures/qualitative_results_on_3dhp.pdf}}
% \caption{Qualitative results on MPI-INF-3DHP.
% % The blue pose represents the predicted results, while the red pose represents the ground truth.
% }
% \label{fig:qualitative_vis_3dhp}
% \end{figure}

% \begin{figure}[htbp]
% \centering
% \centerline{\includegraphics[width=1.0\linewidth]{figures/Frame26.pdf}}
% \caption{
% Qualitative results on Animal3D and CtrlAni3D. 
% % Quantitative results on Animal3D (left column) and CtrlAni3D (right column) dataset. 
% % Blue represents the predicted results, and red represents the ground truth.
% }
% \label{fig:vis_animal3d}
% \end{figure}

\begin{figure*}[htbp]
\centering
\centerline{\includegraphics[width=1.0\linewidth]{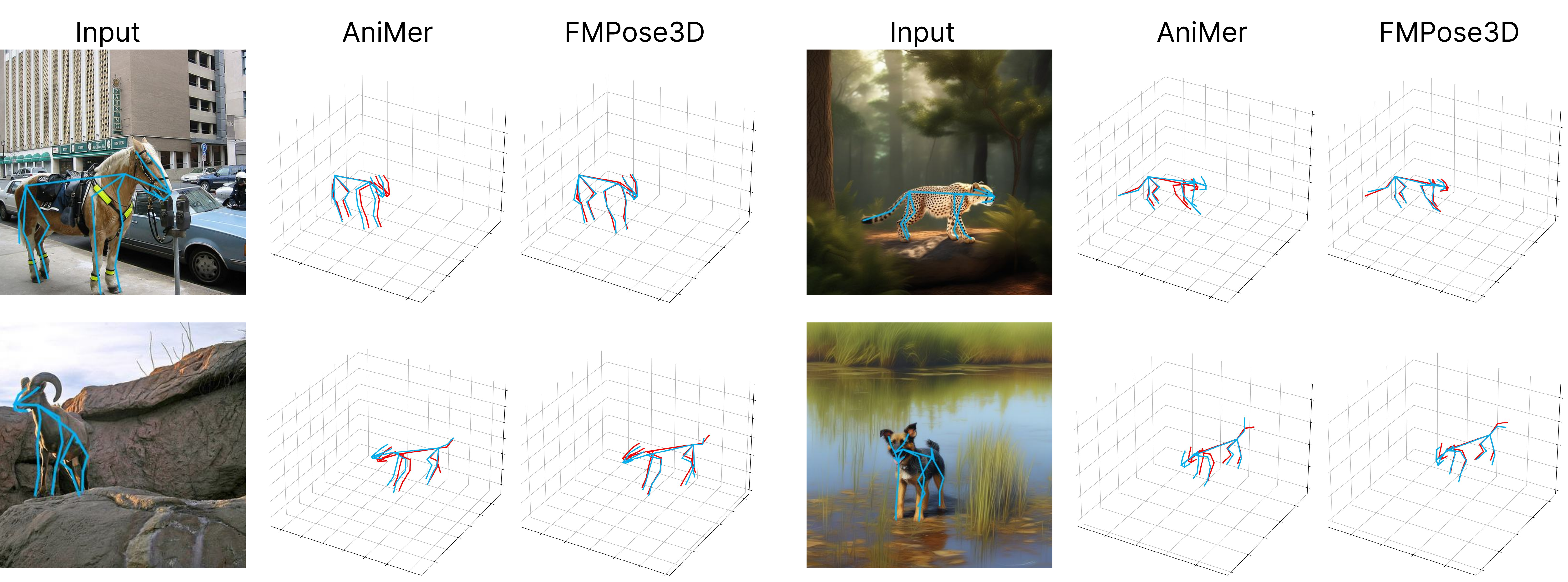}}
\caption{
Qualitative comparison of AniMer \cite{lyu2025animer} and FMPose3D on Animal3D (left column) and CtrlAni3D (right column). 
% Blue represents the predicted results, and red represents the ground truth.
}
\label{fig:vis_animal3d}
\end{figure*}

\subsection{Inference Speed}
Table~\ref{tab:inference_speed} reports the inference speed measured on a single GeForce RTX 4090 GPU.
With $S=3$ ODE steps, FMPose3D runs at 160.11 FPS for a single hypothesis ($N=1$).
Thanks to its parallel design, generating 40 hypotheses still achieves 145.59 FPS with only a minor drop.
To compare our method with diffusion-based models, we reproduce DiffPose~\cite{gong2023diffpose} and measure its FPS in a single-frame setting. Without using DDIM, DiffPose attains only 3.36 FPS, while using DDIM with 5 sampling steps improves the speed to 27.15 FPS. These results show that FMPose3D is substantially faster: even with $N=40$ hypotheses, FMPose3D is still about $5.4\times$ faster than DiffPose.

For additional results on uncertainty estimation see Suppl. Sec. \ref{sec:supp_uncertainty}, and for the effect of training set size Suppl. Sec. \ref{sec:supp_effect_training_set_size}.

\subsection{Qualitative Results}
Figure~\ref{fig:qualitative_vis_h36m} compares our method with DiffPose \cite{gong2023diffpose} on Human3.6M. Across diverse poses, our model produces more accurate and plausible 3D poses. 
% For instance, in the sitting example (column 1, row 2), the two legs in DiffPose’s prediction deviate from the ground truth, whereas the leg angles in our prediction are aligned with the ground truth.
% Additional qualitative results on MPI-INF-3DHP are provided in the Supplementary Material.
% Results on MPI-INF-3DHP (Figure~\ref{fig:qualitative_vis_3dhp}) further show consistent performance in both indoor and outdoor scenes.
Figure~\ref{fig:vis_animal3d} shows the qualitative results on animal datasets. Our FMPose3D successfully estimates the 3D pose for both real (Animal3D, left column) and synthetic (CtrlAni3D, right column) inputs. Our method demonstrates that, despite variations in species, poses, and viewing angles, it consistently produces accurate 3D pose estimates.
For more qualitative results, see Suppl. Sec. \ref{sec:suppl_qualitative}.
\section{Conclusions}

We present \textbf{FMPose3D}, a flow-matching–based generative framework for monocular 3D pose estimation. 
By learning a conditional ODE-driven velocity field, FMPose3D transports samples from Gaussian noise to the distribution of plausible 3D poses conditioned on 2D inputs.
This formulation enables fast inference with only a few ODE integration steps and naturally supports multi-hypothesis generation through different noise initializations.
To exploit this generative capability and obtain a single accurate prediction, we further propose Reprojection-based Posterior Expectation Aggregation (RPEA), which approximates the Bayesian posterior using re-projection error as a likelihood proxy.
Experimental results on both human and animal datasets validate the superior performance of FMPose3D across domains.

\section*{Acknowledgments}

We thank the Swiss National Science Foundation Grant No. 320030-227871 and the Kavli Foundation for funding. We thank members of the M-Lab of Adaptive Intelligence for comments throughout the project.
% \input{sec/X_suppl}
%\clearpage

{
    \small
    \bibliographystyle{ieeenat_fullname}
    \bibliography{main}

\begin{thebibliography}{77}
\providecommand{\natexlab}[1]{#1}
\providecommand{\url}[1]{\texttt{#1}}
\expandafter\ifx\csname urlstyle\endcsname\relax
  \providecommand{\doi}[1]{doi: #1}\else
  \providecommand{\doi}{doi: \begingroup \urlstyle{rm}\Url}\fi

\bibitem[Azizi et~al.(2024)Azizi, Fayyaz, and Bischof]{azizi2024occlusion}
Niloofar Azizi, Mohsen Fayyaz, and Horst Bischof.
\newblock Occlusion handling in 3d human pose estimation with perturbed
  positional encoding.
\newblock In \emph{ECCV}, pages 441--458, 2024.

\bibitem[Ba et~al.(2016)Ba, Kiros, and Hinton]{ba2016layer}
Jimmy~Lei Ba, Jamie~Ryan Kiros, and Geoffrey~E Hinton.
\newblock Layer normalization.
\newblock \emph{arXiv preprint arXiv:1607.06450}, 2016.

\bibitem[Baradel et~al.(2024)Baradel, Armando, Galaaoui, Br{\'e}gier,
  Weinzaepfel, Rogez, and Lucas]{baradel2024multi}
Fabien Baradel, Matthieu Armando, Salma Galaaoui, Romain Br{\'e}gier, Philippe
  Weinzaepfel, Gr{\'e}gory Rogez, and Thomas Lucas.
\newblock Multi-hmr: Multi-person whole-body human mesh recovery in a single
  shot.
\newblock In \emph{ECCV}, pages 202--218, 2024.

\bibitem[Berger(2013)]{berger2013statistical}
James~O Berger.
\newblock \emph{Statistical decision theory and Bayesian analysis}.
\newblock 2013.

\bibitem[Biggs et~al.(2020)Biggs, Boyne, Charles, Fitzgibbon, and
  Cipolla]{biggs2020left}
Benjamin Biggs, Oliver Boyne, James Charles, Andrew Fitzgibbon, and Roberto
  Cipolla.
\newblock Who left the dogs out? 3d animal reconstruction with expectation
  maximization in the loop.
\newblock In \emph{ECCV}, pages 195--211, 2020.

\bibitem[Bishop(1994)]{bishop1994mixture}
Christopher~M Bishop.
\newblock Mixture density networks.
\newblock 1994.

\bibitem[Cai et~al.(2019)Cai, Ge, Liu, Cai, Cham, Yuan, and
  Thalmann]{cai2019exploiting}
Yujun Cai, Liuhao Ge, Jun Liu, Jianfei Cai, Tat-Jen Cham, Junsong Yuan, and
  Nadia~Magnenat Thalmann.
\newblock Exploiting spatial-temporal relationships for {3D} pose estimation
  via graph convolutional networks.
\newblock In \emph{ICCV}, pages 2272--2281, 2019.

\bibitem[Chen et~al.(2018)Chen, Wang, Peng, Zhang, Yu, and
  Sun]{chen2018cascaded}
Yilun Chen, Zhicheng Wang, Yuxiang Peng, Zhiqiang Zhang, Gang Yu, and Jian Sun.
\newblock Cascaded pyramid network for multi-person pose estimation.
\newblock In \emph{CVPR}, pages 7103--7112, 2018.

\bibitem[Ci et~al.(2019)Ci, Wang, Ma, and Wang]{ci2019optimizing}
Hai Ci, Chunyu Wang, Xiaoxuan Ma, and Yizhou Wang.
\newblock Optimizing network structure for {3D} human pose estimation.
\newblock In \emph{ICCV}, pages 2262--2271, 2019.

\bibitem[Deng et~al.(2009)Deng, Dong, Socher, Li, Li, and
  Fei-Fei]{deng2009imagenet}
Jia Deng, Wei Dong, Richard Socher, Li-Jia Li, Kai Li, and Li Fei-Fei.
\newblock Imagenet: A large-scale hierarchical image database.
\newblock In \emph{CVPR}, pages 248--255, 2009.

\bibitem[Dhariwal and Nichol(2021)]{dhariwal2021diffusion}
Prafulla Dhariwal and Alexander Nichol.
\newblock Diffusion models beat gans on image synthesis.
\newblock In \emph{NeurIPS}, pages 8780--8794, 2021.

\bibitem[Fei et~al.(2024)Fei, Wu, Ji, Zhang, and Chua]{fei2024dysen}
Hao Fei, Shengqiong Wu, Wei Ji, Hanwang Zhang, and Tat-Seng Chua.
\newblock Dysen-vdm: Empowering dynamics-aware text-to-video diffusion with
  llms.
\newblock In \emph{CVPR}, pages 7641--7653, 2024.

\bibitem[Gholami et~al.(2022)Gholami, Wandt, Rhodin, Ward, and
  Wang]{gholami2022adaptpose}
Mohsen Gholami, Bastian Wandt, Helge Rhodin, Rabab Ward, and Z~Jane Wang.
\newblock Adaptpose: Cross-dataset adaptation for {3D} human pose estimation by
  learnable motion generation.
\newblock In \emph{CVPR}, pages 13075--13085, 2022.

\bibitem[Glorot et~al.(2011)Glorot, Bordes, and Bengio]{glorot2011deep}
Xavier Glorot, Antoine Bordes, and Yoshua Bengio.
\newblock Deep sparse rectifier neural networks.
\newblock In \emph{Proceedings of the fourteenth international conference on
  artificial intelligence and statistics}, pages 315--323, 2011.

\bibitem[Goel et~al.(2023)Goel, Pavlakos, Rajasegaran, Kanazawa, and
  Malik]{goel2023humans}
Shubham Goel, Georgios Pavlakos, Jathushan Rajasegaran, Angjoo Kanazawa, and
  Jitendra Malik.
\newblock Humans in 4d: Reconstructing and tracking humans with transformers.
\newblock In \emph{ICCV}, pages 14783--14794, 2023.

\bibitem[Gong et~al.(2023)Gong, Foo, Fan, Ke, Rahmani, and
  Liu]{gong2023diffpose}
Jia Gong, Lin~Geng Foo, Zhipeng Fan, Qiuhong Ke, Hossein Rahmani, and Jun Liu.
\newblock Diffpose: Toward more reliable {3D} pose estimation.
\newblock In \emph{CVPR}, pages 13041--13051, 2023.

\bibitem[Goodfellow et~al.(2020)Goodfellow, Pouget-Abadie, Mirza, Xu,
  Warde-Farley, Ozair, Courville, and Bengio]{goodfellow2020generative}
Ian Goodfellow, Jean Pouget-Abadie, Mehdi Mirza, Bing Xu, David Warde-Farley,
  Sherjil Ozair, Aaron Courville, and Yoshua Bengio.
\newblock Generative adversarial networks.
\newblock \emph{Communications of the ACM}, 63\penalty0 (11):\penalty0
  139--144, 2020.

\bibitem[Gui et~al.(2025)Gui, Schusterbauer, Prestel, Ma, Kotovenko,
  Grebenkova, Baumann, Hu, and Ommer]{gui2025depthfm}
Ming Gui, Johannes Schusterbauer, Ulrich Prestel, Pingchuan Ma, Dmytro
  Kotovenko, Olga Grebenkova, Stefan~Andreas Baumann, Vincent~Tao Hu, and
  Bj{\"o}rn Ommer.
\newblock Depthfm: Fast generative monocular depth estimation with flow
  matching.
\newblock In \emph{AAAI}, pages 3203--3211, 2025.

\bibitem[Han et~al.(2025)Han, Kim, and Lee]{han2025propose}
Jumin Han, Jun-Hee Kim, and Seong-Whan Lee.
\newblock Propose: Probabilistic 3d human pose estimation with instance-level
  distribution and normalizing flow.
\newblock In \emph{AAAI}, pages 3338--3346, 2025.

\bibitem[Ho et~al.(2020)Ho, Jain, and Abbeel]{ho2020denoising}
Jonathan Ho, Ajay Jain, and Pieter Abbeel.
\newblock Denoising diffusion probabilistic models.
\newblock In \emph{NeurIPS}, pages 6840--6851, 2020.

\bibitem[Holmquist and Wandt(2023)]{holmquist2023diffpose}
Karl Holmquist and Bastian Wandt.
\newblock Diffpose: Multi-hypothesis human pose estimation using diffusion
  models.
\newblock In \emph{ICCV}, pages 15977--15987, 2023.

\bibitem[Hu et~al.(2023)Hu, Liu, Li, Yan, Fang, and Chen]{hu2023geometric}
Mengxian Hu, Chengju Liu, Shu Li, Qingqing Yan, Qin Fang, and Qijun Chen.
\newblock A geometric knowledge oriented single-frame {2D}-to-{3D} human
  absolute pose estimation method.
\newblock \emph{IEEE Transactions on Circuits and Systems for Video
  Technology}, 2023.

\bibitem[Huang et~al.(2023)Huang, Park, Wang, Denk, Ly, Chen, Zhang, Zhang, Yu,
  Frank, et~al.]{huang2023noise2music}
Qingqing Huang, Daniel~S Park, Tao Wang, Timo~I Denk, Andy Ly, Nanxin Chen,
  Zhengdong Zhang, Zhishuai Zhang, Jiahui Yu, Christian Frank, et~al.
\newblock Noise2music: Text-conditioned music generation with diffusion models.
\newblock \emph{arXiv preprint arXiv:2302.03917}, 2023.

\bibitem[Ionescu et~al.(2013)Ionescu, Papava, Olaru, and Sminchisescu]{h36m}
Catalin Ionescu, Dragos Papava, Vlad Olaru, and Cristian Sminchisescu.
\newblock Human3.6m: Large scale datasets and predictive methods for {3D} human
  sensing in natural environments.
\newblock \emph{IEEE Transactions on Pattern Analysis and Machine
  Intelligence}, 36\penalty0 (7):\penalty0 1325--1339, 2013.

\bibitem[Jahangiri and Yuille(2017)]{jahangiri2017generating}
Ehsan Jahangiri and Alan~L Yuille.
\newblock Generating multiple diverse hypotheses for human 3d pose consistent
  with 2d joint detections.
\newblock In \emph{ICCV Workshops}, pages 805--814, 2017.

\bibitem[Jiang et~al.(2022)Jiang, Lee, Teotia, and Ostadabbas]{jiang2022animal}
Le Jiang, Caleb Lee, Divyang Teotia, and Sarah Ostadabbas.
\newblock Animal pose estimation: A closer look at the state-of-the-art,
  existing gaps and opportunities.
\newblock \emph{Computer Vision and Image Understanding}, 222:\penalty0 103483,
  2022.

\bibitem[Jiang et~al.(2024)Jiang, Zhou, Li, Chai, Yang, and
  Hwang]{jiang2024back}
Zhongyu Jiang, Zhuoran Zhou, Lei Li, Wenhao Chai, Cheng-Yen Yang, and Jenq-Neng
  Hwang.
\newblock Back to optimization: Diffusion-based zero-shot 3d human pose
  estimation.
\newblock In \emph{WACV}, pages 6142--6152, 2024.

\bibitem[Kanazawa et~al.(2018)Kanazawa, Black, Jacobs, and
  Malik]{kanazawa2018end}
Angjoo Kanazawa, Michael~J Black, David~W Jacobs, and Jitendra Malik.
\newblock End-to-end recovery of human shape and pose.
\newblock In \emph{CVPR}, pages 7122--7131, 2018.

\bibitem[Kaye et~al.(2025)Kaye, Jakab, Wu, Ruprecht, and
  Vedaldi]{kaye2025dualpm}
Ben Kaye, Tomas Jakab, Shangzhe Wu, Christian Ruprecht, and Andrea Vedaldi.
\newblock Dualpm: dual posed-canonical point maps for 3d shape and pose
  reconstruction.
\newblock In \emph{CVPR}, pages 6425--6435, 2025.

\bibitem[Kipf and Welling(2016)]{kipf2016semi}
Thomas~N Kipf and Max Welling.
\newblock Semi-supervised classification with graph convolutional networks.
\newblock \emph{arXiv preprint arXiv:1609.02907}, 2016.

\bibitem[Kipf and Welling(2017)]{kipf2017semisupervised}
Thomas~N. Kipf and Max Welling.
\newblock Semi-supervised classification with graph convolutional networks.
\newblock In \emph{International Conference on Learning Representations}, 2017.

\bibitem[Li and Lee(2019)]{li2019generating}
Chen Li and Gim~Hee Lee.
\newblock Generating multiple hypotheses for 3d human pose estimation with
  mixture density network.
\newblock In \emph{CVPR}, pages 9887--9895, 2019.

\bibitem[Li and Lee(2020)]{li2020weakly}
Chen Li and Gim~Hee Lee.
\newblock Weakly supervised generative network for multiple 3d human pose
  hypotheses.
\newblock In \emph{BMVC}, 2020.

\bibitem[Li et~al.(2023)Li, Shi, Dai, Zheng, Wang, Sun, Guo, Li, Zou, and
  Xiong]{li2023pose}
Han Li, Bowen Shi, Wenrui Dai, Hongwei Zheng, Botao Wang, Yu Sun, Min Guo,
  Chenglin Li, Junni Zou, and Hongkai Xiong.
\newblock Pose-oriented transformer with uncertainty-guided refinement for
  {2D}-to-{3D} human pose estimation.
\newblock In \emph{AAAI}, pages 1296--1304, 2023.

\bibitem[Li et~al.(2022{\natexlab{a}})Li, Liu, Ding, Liu, Wang, and
  Yang]{strided}
Wenhao Li, Hong Liu, Runwei Ding, Mengyuan Liu, Pichao Wang, and Wenming Yang.
\newblock Exploiting temporal contexts with strided transformer for {3D} human
  pose estimation.
\newblock \emph{IEEE Transactions on Multimedia}, pages 1282--1293,
  2022{\natexlab{a}}.

\bibitem[Li et~al.(2022{\natexlab{b}})Li, Liu, Tang, Wang, and
  Van~Gool]{li2022mhformer}
Wenhao Li, Hong Liu, Hao Tang, Pichao Wang, and Luc Van~Gool.
\newblock {MHF}ormer: Multi-hypothesis transformer for {3D} human pose
  estimation.
\newblock In \emph{CVPR}, pages 13147--13156, 2022{\natexlab{b}}.

\bibitem[Li et~al.(2024)Li, Litvak, Li, Zhang, Jakab, Rupprecht, Wu, Vedaldi,
  and Wu]{li2024learning}
Zizhang Li, Dor Litvak, Ruining Li, Yunzhi Zhang, Tomas Jakab, Christian
  Rupprecht, Shangzhe Wu, Andrea Vedaldi, and Jiajun Wu.
\newblock Learning the 3d fauna of the web.
\newblock In \emph{CVPR}, pages 9752--9762, 2024.

\bibitem[Lin et~al.(2014)Lin, Maire, Belongie, Hays, Perona, Ramanan,
  Doll{\'a}r, and Zitnick]{lin2014microsoftcoco}
Tsung-Yi Lin, Michael Maire, Serge Belongie, James Hays, Pietro Perona, Deva
  Ramanan, Piotr Doll{\'a}r, and C~Lawrence Zitnick.
\newblock Microsoft coco: Common objects in context.
\newblock In \emph{ECCV}, pages 740--755, 2014.

\bibitem[Lipman et~al.(2023)Lipman, Chen, Ben-Hamu, Nickel, and
  Le]{lipman2023flow}
Yaron Lipman, Ricky~TQ Chen, Heli Ben-Hamu, Maximilian Nickel, and Matt Le.
\newblock Flow matching for generative modeling.
\newblock In \emph{ICLR}, 2023.

\bibitem[Liu et~al.(2023{\natexlab{a}})Liu, Chen, Yuan, Mei, Liu, Mandic, Wang,
  and Plumbley]{liu2023audioldm}
Haohe Liu, Zehua Chen, Yi Yuan, Xinhao Mei, Xubo Liu, Danilo Mandic, Wenwu
  Wang, and Mark~D Plumbley.
\newblock Audioldm: Text-to-audio generation with latent diffusion models.
\newblock \emph{arXiv preprint arXiv:2301.12503}, 2023{\natexlab{a}}.

\bibitem[Liu et~al.(2022)Liu, Bao, Sun, and Mei]{liu2022recent}
Wu Liu, Qian Bao, Yu Sun, and Tao Mei.
\newblock Recent advances of monocular 2d and 3d human pose estimation: A deep
  learning perspective.
\newblock \emph{ACM Computing Surveys}, 55\penalty0 (4):\penalty0 1--41, 2022.

\bibitem[Liu et~al.(2023{\natexlab{b}})Liu, Gong, and
  Liu]{liu2023rectifiedflow}
Xingchao Liu, Chengyue Gong, and Qiang Liu.
\newblock Flow straight and fast: Learning to generate and transfer data with
  rectified flow.
\newblock In \emph{ICLR}, 2023{\natexlab{b}}.

\bibitem[Lyu et~al.(2025)Lyu, Zhu, Gu, Lin, Cheng, Liu, Tang, and
  An]{lyu2025animer}
Jin Lyu, Tianyi Zhu, Yi Gu, Li Lin, Pujin Cheng, Yebin Liu, Xiaoying Tang, and
  Liang An.
\newblock Animer: Animal pose and shape estimation using family aware
  transformer.
\newblock In \emph{CVPR}, pages 17486--17496, 2025.

\bibitem[Martinez et~al.(2017)Martinez, Hossain, Romero, and
  Little]{martinez2017simple}
Julieta Martinez, Rayat Hossain, Javier Romero, and James~J Little.
\newblock A simple yet effective baseline for {3D} human pose estimation.
\newblock In \emph{ICCV}, pages 2640--2649, 2017.

\bibitem[Mehta et~al.(2017)Mehta, Rhodin, Casas, Fua, Sotnychenko, Xu, and
  Theobalt]{mehta2017monocular}
Dushyant Mehta, Helge Rhodin, Dan Casas, Pascal Fua, Oleksandr Sotnychenko,
  Weipeng Xu, and Christian Theobalt.
\newblock Monocular {3D} human pose estimation in the wild using improved cnn
  supervision.
\newblock In \emph{3DV}, pages 506--516, 2017.

\bibitem[Oikarinen et~al.(2021)Oikarinen, Hannah, and
  Kazerounian]{oikarinen2021graphmdn}
Tuomas Oikarinen, Daniel Hannah, and Sohrob Kazerounian.
\newblock Graphmdn: Leveraging graph structure and deep learning to solve
  inverse problems.
\newblock In \emph{IJCNN}, pages 1--9, 2021.

\bibitem[Pavllo et~al.(2019)Pavllo, Feichtenhofer, Grangier, and
  Auli]{pavllo20193d}
Dario Pavllo, Christoph Feichtenhofer, David Grangier, and Michael Auli.
\newblock {3D} human pose estimation in video with temporal convolutions and
  semi-supervised training.
\newblock In \emph{CVPR}, pages 7753--7762, 2019.

\bibitem[Ramesh et~al.(2022)Ramesh, Dhariwal, Nichol, Chu, and
  Chen]{ramesh2022hierarchical}
Aditya Ramesh, Prafulla Dhariwal, Alex Nichol, Casey Chu, and Mark Chen.
\newblock Hierarchical text-conditional image generation with clip latents.
\newblock \emph{arXiv preprint arXiv:2204.06125}, 1\penalty0 (2):\penalty0 3,
  2022.

\bibitem[Shan et~al.(2023)Shan, Liu, Zhang, Wang, Han, Wang, Ma, and
  Gao]{shan2023diffusion}
Wenkang Shan, Zhenhua Liu, Xinfeng Zhang, Zhao Wang, Kai Han, Shanshe Wang,
  Siwei Ma, and Wen Gao.
\newblock Diffusion-based 3d human pose estimation with multi-hypothesis
  aggregation.
\newblock In \emph{ICCV}, pages 14761--14771, 2023.

\bibitem[Sharma et~al.(2019)Sharma, Varigonda, Bindal, Sharma, and
  Jain]{sharma2019monocular}
Saurabh Sharma, Pavan~Teja Varigonda, Prashast Bindal, Abhishek Sharma, and
  Arjun Jain.
\newblock Monocular 3d human pose estimation by generation and ordinal ranking.
\newblock In \emph{ICCV}, pages 2325--2334, 2019.

\bibitem[Sohn et~al.(2015)Sohn, Lee, and Yan]{sohn2015learning}
Kihyuk Sohn, Honglak Lee, and Xinchen Yan.
\newblock Learning structured output representation using deep conditional
  generative models.
\newblock In \emph{NeurIPS}, 2015.

\bibitem[Song et~al.(2021)Song, Meng, and Ermon]{song2021denoising}
Jiaming Song, Chenlin Meng, and Stefano Ermon.
\newblock Denoising diffusion implicit models.
\newblock In \emph{ICLR}, 2021.

\bibitem[Stathopoulos et~al.(2024)Stathopoulos, Han, and
  Metaxas]{stathopoulos2024score}
Anastasis Stathopoulos, Ligong Han, and Dimitris Metaxas.
\newblock Score-guided diffusion for 3d human recovery.
\newblock In \emph{CVPR}, pages 906--915, 2024.

\bibitem[Sun et~al.(2019)Sun, Xiao, Liu, and Wang]{sun2019deep}
Ke Sun, Bin Xiao, Dong Liu, and Jingdong Wang.
\newblock Deep high-resolution representation learning for human pose
  estimation.
\newblock In \emph{CVPR}, pages 5693--5703, 2019.

\bibitem[Tang et~al.(2023)Tang, Li, Hao, and Hong]{tang2023mlp}
Zhenhua Tang, Jia Li, Yanbin Hao, and Richang Hong.
\newblock Mlp-jcg: Multi-layer perceptron with joint-coordinate gating for
  efficient 3d human pose estimation.
\newblock \emph{IEEE Transactions on Multimedia}, 2023.

\bibitem[Vaswani et~al.(2017)Vaswani, Shazeer, Parmar, Uszkoreit, Jones, Gomez,
  Kaiser, and Polosukhin]{Attention}
Ashish Vaswani, Noam Shazeer, Niki Parmar, Jakob Uszkoreit, Llion Jones,
  Aidan~N Gomez, {\L}ukasz Kaiser, and Illia Polosukhin.
\newblock Attention is all you need.
\newblock In \emph{NeurIPS}, pages 5998--6008, 2017.

\bibitem[von Marcard et~al.(2018)von Marcard, Henschel, Black, Rosenhahn, and
  Pons-Moll]{pw3d}
Timo von Marcard, Roberto Henschel, Michael~J Black, Bodo Rosenhahn, and Gerard
  Pons-Moll.
\newblock Recovering accurate {3D} human pose in the wild using {IMUs} and a
  moving camera.
\newblock In \emph{ECCV}, pages 601--617, 2018.

\bibitem[Wang et~al.(2024{\natexlab{a}})Wang, Li, Qi, Ding, Tong, and
  Yang]{wang2024semflow}
Chaoyang Wang, Xiangtai Li, Lu Qi, Henghui Ding, Yunhai Tong, and Ming-Hsuan
  Yang.
\newblock Semflow: Binding semantic segmentation and image synthesis via
  rectified flow.
\newblock In \emph{NeurIPS}, pages 138981--139001, 2024{\natexlab{a}}.

\bibitem[Wang et~al.(2020)Wang, Sun, Cheng, Jiang, Deng, Zhao, Liu, Mu, Tan,
  Wang, et~al.]{wang2020deep}
Jingdong Wang, Ke Sun, Tianheng Cheng, Borui Jiang, Chaorui Deng, Yang Zhao,
  Dong Liu, Yadong Mu, Mingkui Tan, Xinggang Wang, et~al.
\newblock Deep high-resolution representation learning for visual recognition.
\newblock \emph{TPAMI}, 43\penalty0 (10):\penalty0 3349--3364, 2020.

\bibitem[Wang et~al.(2025)Wang, Lin, Sun, Liu, Nie, Li, Liao, Chu, and
  Zhao]{wang2025editor}
JiYuan Wang, Chunyu Lin, Lei Sun, Rongying Liu, Lang Nie, Mingxing Li, Kang
  Liao, Xiangxiang Chu, and Yao Zhao.
\newblock From editor to dense geometry estimator.
\newblock \emph{arXiv preprint arXiv:2509.04338}, 2025.

\bibitem[Wang et~al.(2024{\natexlab{b}})Wang, Xiao, Wang, Liu, Wang, and
  Chen]{wang2024text}
Weiquan Wang, Jun Xiao, Chunping Wang, Wei Liu, Zhao Wang, and Long Chen.
\newblock Di\textsuperscript{2}pose: Discrete diffusion model for occluded 3d
  human pose estimation.
\newblock In \emph{NeurIPS}, pages 98717--98741, 2024{\natexlab{b}}.

\bibitem[Wehrbein et~al.(2021)Wehrbein, Rudolph, Rosenhahn, and
  Wandt]{wehrbein2021probabilistic}
Tom Wehrbein, Marco Rudolph, Bodo Rosenhahn, and Bastian Wandt.
\newblock Probabilistic monocular 3d human pose estimation with normalizing
  flows.
\newblock In \emph{ICCV}, pages 11199--11208, 2021.

\bibitem[Xu et~al.(2023)Xu, Zhang, Peng, Ma, Jesslen, Ji, Hu, Zhang, Liu, Wang,
  et~al.]{xu2023animal3d}
Jiacong Xu, Yi Zhang, Jiawei Peng, Wufei Ma, Artur Jesslen, Pengliang Ji, Qixin
  Hu, Jiehua Zhang, Qihao Liu, Jiahao Wang, et~al.
\newblock Animal3d: A comprehensive dataset of 3d animal pose and shape.
\newblock In \emph{ICCV}, pages 9099--9109, 2023.

\bibitem[Xu and Takano(2021)]{xu2021graph}
Tianhan Xu and Wataru Takano.
\newblock Graph stacked hourglass networks for {3D} human pose estimation.
\newblock In \emph{CVPR}, pages 16105--16114, 2021.

\bibitem[You et~al.(2023)You, Liu, Wang, Li, Ding, and Li]{you2023co}
Yingxuan You, Hong Liu, Ti Wang, Wenhao Li, Runwei Ding, and Xia Li.
\newblock Co-evolution of pose and mesh for 3d human body estimation from
  video.
\newblock In \emph{ICCV}, pages 14963--14973, 2023.

\bibitem[Zeng et~al.(2020)Zeng, Sun, Huang, Liu, Xu, and Lin]{zeng2020srnet}
Ailing Zeng, Xiao Sun, Fuyang Huang, Minhao Liu, Qiang Xu, and Stephen Lin.
\newblock {SRN}et: Improving generalization in {3D} human pose estimation with
  a split-and-recombine approach.
\newblock In \emph{ECCV}, pages 507--523, 2020.

\bibitem[Zeng et~al.(2021)Zeng, Sun, Yang, Zhao, Liu, and Xu]{zeng2021learning}
Ailing Zeng, Xiao Sun, Lei Yang, Nanxuan Zhao, Minhao Liu, and Qiang Xu.
\newblock Learning skeletal graph neural networks for hard {3D} pose
  estimation.
\newblock In \emph{ICCV}, pages 11436--11445, 2021.

\bibitem[Zhang and Carlone(2025)]{zhang2025champ}
Harry Zhang and Luca Carlone.
\newblock Champ: Conformalized 3d human multi-hypothesis pose estimators.
\newblock In \emph{ICLR}, 2025.

\bibitem[Zhang et~al.(2022)Zhang, Tu, Yang, Chen, and Yuan]{zhang2022mixste}
Jinlu Zhang, Zhigang Tu, Jianyu Yang, Yujin Chen, and Junsong Yuan.
\newblock Mix{STE}: Seq2seq mixed spatio-temporal encoder for {3D} human pose
  estimation in video.
\newblock In \emph{CVPR}, pages 13232--13242, 2022.

\bibitem[Zhang et~al.(2023)Zhang, Rao, and Agrawala]{zhang2023adding}
Lvmin Zhang, Anyi Rao, and Maneesh Agrawala.
\newblock Adding conditional control to text-to-image diffusion models.
\newblock In \emph{ICCV}, pages 3836--3847, 2023.

\bibitem[Zhao et~al.(2019)Zhao, Peng, Tian, Kapadia, and
  Metaxas]{zhao2019semantic}
Long Zhao, Xi Peng, Yu Tian, Mubbasir Kapadia, and Dimitris~N Metaxas.
\newblock Semantic graph convolutional networks for {3D} human pose regression.
\newblock In \emph{CVPR}, pages 3425--3435, 2019.

\bibitem[Zhao et~al.(2022)Zhao, Wang, and Tian]{zhao2022graformer}
Weixi Zhao, Weiqiang Wang, and Yunjie Tian.
\newblock Gra{F}ormer: Graph-oriented transformer for {3D} pose estimation.
\newblock In \emph{CVPR}, pages 20438--20447, 2022.

\bibitem[Zheng et~al.(2021)Zheng, Zhu, Mendieta, Yang, Chen, and
  Ding]{zheng20213d}
Ce Zheng, Sijie Zhu, Matias Mendieta, Taojiannan Yang, Chen Chen, and Zhengming
  Ding.
\newblock {3D} human pose estimation with spatial and temporal transformers.
\newblock In \emph{ICCV}, pages 11656--11665, 2021.

\bibitem[Zheng et~al.(2023)Zheng, Wu, Chen, Yang, Zhu, Shen, Kehtarnavaz, and
  Shah]{zheng2023deep}
Ce Zheng, Wenhan Wu, Chen Chen, Taojiannan Yang, Sijie Zhu, Ju Shen, Nasser
  Kehtarnavaz, and Mubarak Shah.
\newblock Deep learning-based human pose estimation: A survey.
\newblock \emph{ACM computing surveys}, 56\penalty0 (1):\penalty0 1--37, 2023.

\bibitem[Zou and Tang(2021)]{zou2021modulated}
Zhiming Zou and Wei Tang.
\newblock Modulated graph convolutional network for 3d human pose estimation.
\newblock In \emph{ICCV}, pages 11477--11487, 2021.

\bibitem[Zuffi et~al.(2017)Zuffi, Kanazawa, Jacobs, and Black]{zuffi20173d}
Silvia Zuffi, Angjoo Kanazawa, David~W Jacobs, and Michael~J Black.
\newblock 3d menagerie: Modeling the 3d shape and pose of animals.
\newblock In \emph{CVPR}, pages 6365--6373, 2017.

\bibitem[Zuffi et~al.(2019)Zuffi, Kanazawa, Berger-Wolf, and
  Black]{zuffi2019three}
Silvia Zuffi, Angjoo Kanazawa, Tanya Berger-Wolf, and Michael~J Black.
\newblock Three-d safari: Learning to estimate zebra pose, shape, and texture
  from images" in the wild".
\newblock In \emph{ICCV}, pages 5359--5368, 2019.

\end{thebibliography}
}
\renewcommand{\theequation}{\Alph{equation}}
\renewcommand{\thesection}{\Alph{section}}
\renewcommand{\thesubsection}{\Alph{section}.\arabic{subsection}}

\clearpage
\setcounter{page}{1}
\maketitlesupplementary
\appendix

\section{Background on Attention and GCN} 
\label{sec:supp_attn_gcn}
\noindent \textbf{Attention.}
\label{sec:attn}
The input tokens $X {\in} \mathbb{R}^{J \times D}$ are first projected to queries $Q {\in} \mathbb{R}^{J\times d}$, keys $K {\in} \mathbb{R}^{J\times d}$, and values $V {\in} \mathbb{R}^{J\times d}$,
and then $Q, K,V$ are fed to a scaled dot-product attention~\cite{Attention}:
\begin{equation}
    \text{Attention}(Q,K,V)=\text{Softmax}(QK^{T}/\sqrt{d})V,
    \label{equ:MSA}
\end{equation}
where $d$ is the dimension of $Q, K, V$. Multi-head self-attention~(MSA)~\cite{Attention} splits $Q$, $K$, $V$ into multiple heads, each of which applies scaled dot-product attention in parallel.
% The formula can be expressed as:
% \begin{equation}
%     \text{MSA}(Q,K,V)=\text{Concat}(H_1,H_2,...,H_h)W,
%     \label{equ:MSA}
% \end{equation}
% where $H_i {=} \text{Attention}(Q_i,K_i,V_i)$ denotes the $i^{th}{\in} [1,...,h]$ head, and $W$ is the projection transformation matrix.
This enables the model to efficiently utilize information from various representation subspaces with different locations.

\noindent \textbf{Graph Convolutional Network.} 
Graph Convolutional Network (GCN) \cite{kipf2016semi} is capable of capturing intricate relationships and structures within graph-structured data.
Consider an undirected graph $G {=} \{V, E\}$, where $V$ is the set of nodes and $E$ is the set of edges.
The edges can be encoded in an adjacency matrix $A {\in} \{0,1\}^{N {\times} N}$. 
For the input $X_l$ of the $l^{th}$ layer, the vanilla graph convolution aggregates the features of the neighboring nodes. The output  $X_{l+1}$  of the $l^{th}$ GCN layer can be formulated as:
\begin{equation}
    X_{l+1}=\sigma\left(\tilde{D}^{-\frac{1}{2}} \tilde{A} \tilde{D}^{-\frac{1}{2}} X_{l} W \right),
    \label{equ:gcn}
\end{equation}
where $\sigma$ is the ReLU activation function \cite{glorot2011deep}, $W_{l} {\in} \mathbb{R}^{d_{1} \times d} $ is the layer-specific trainable weight matrix.
$\tilde{A} {=} A {+} I_N$ is the adjacency matrix of the graph with added self-connections, where $I_N$ is the identity matrix. Additionally, $\tilde{D}$ is the diagonal node degree matrix.
By stacking multiple GCN layers, it iteratively transforms and aggregates neighboring nodes, thereby obtaining enhanced feature representations.
% For the task of 3D human pose estimation, GCN can effectively capture the semantic structure of the human skeleton.

% \section{Experiments}

\section{Additional Implementation Details}
\label{sec:suppl_implementation}
\noindent \textbf{Humans.}
For Human3.6M \cite{h36m} and MPI-INF-3DHP \cite{mehta2017monocular}, each sample contains $J = 17$ joints. For the model architecture described in Sec.~3.3, each block takes the embedding as input, feeds it into a parallel structure with a GCN branch and an attention branch, concatenates the resulting features, and then processes them with an MLP layer, as illustrated in Figure~2 of the main paper. This block is repeated $L = 5$ times.
The dimensionality of the concatenated feature embedding is set to $D = 512$.  The learning rate is initialized at $0.001$ and is multiplied by $0.98$ at each epoch, with the factor replaced by $0.8$ every 5 epochs. During inference,  we set the number of ODE integration steps to $S = 3$. We employ horizontal flip augmentation, following prior works \cite{cai2019exploiting, pavllo20193d, zeng2021learning}. Rather than averaging the predictions from the original and flipped inputs, we treat them as two separate hypotheses and feed both into our RPEA module to obtain the final 3D pose.  We refer to this strategy as \textbf{Flipped Hypothesis Aggregation (FHA)}. Following common practice \cite{xu2021graph,zeng2021learning,cai2019exploiting,pavllo20193d}, we use 2D poses detected by the cascaded pyramid network (CPN) \cite{chen2018cascaded} for Human3.6M,
and the dataset-provided 2D poses for MPI-INF-3DHP.

\noindent \textbf{Animals.}
For Animal3D \cite{xu2023animal3d}  and CtrlAni3D \cite{lyu2025animer}, each sample contains $J = 26$ joints. We train a single model jointly on the two datasets and evaluate it on each dataset individually. The network architecture follows the same design as in the human setting, with the block repeated for $L = 5$ layers. The dimensionality of the concatenated feature embedding is set to $D = 512$. The model is trained for 300 epochs with a batch size of $13$. The learning rate is initialized at $0.001$ and is multiplied by $0.95$ at each epoch, with the factor replaced by $0.75$ every 15 epochs. During inference, we set the number of ODE integration steps to $S = 3$. For the results reported in Table~3 of the main paper, we generate a single prediction per input and do not use flip augmentation or any multi-hypothesis strategy.

\begin{figure}[b]
    \centering
\includegraphics[width=1.0\linewidth]{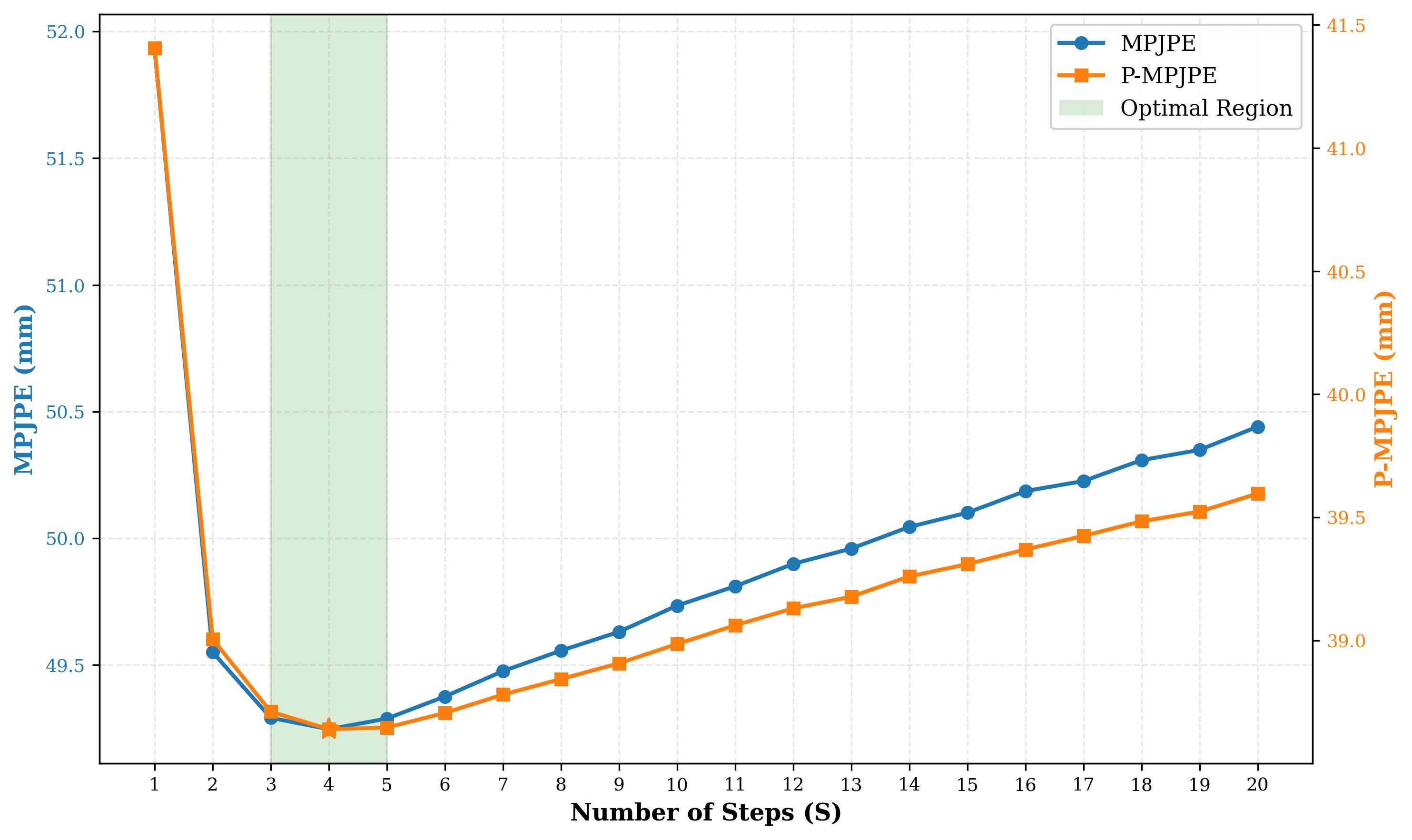}
\caption{
Effect of the number of integration steps $S$ on inference accuracy.
The blue curve shows MPJPE (read from the left vertical axis), and the orange curve shows P-MPJPE (read from the right vertical axis); the shaded region marks the range $S \in \{3,4,5\}$ where both metrics attain their optimal or near-optimal values.
}   
\label{fig:impact_steps}
\end{figure}

\begin{table*}[t]
% \footnotesize
% \setlength\tabcolsep{2mm}
\caption{Quantitative comparison with the state-of-the-art methods on Human3.6M under P-MPJPE. The detected 2D pose is used as input. $N$ denotes the number of hypotheses.
\textcolor{red}{Red}: Best. \textcolor{blue}{Blue}: Second Best.
\colorbox{grey!10}{Grey}: our method.
}
\resizebox{\textwidth}{!}{
\begin{tabular}{lr|ccccccccccccccc|c}
\toprule
\multicolumn{2}{c|}{\textbf{Deterministic Method}} & Dire. & Disc. & Eat & Greet & Phone & Photo & Pose & Purch. & Sit & SitD. & Smoke & Wait & WalkD. & Walk & WalkT. & Avg $\downarrow$  \\
\midrule
SimpleBaseline \cite{martinez2017simple} & ICCV'17 & 39.5 & 43.2 & 46.4 & 47.0 & 51.0 & 56.0 & 41.4 & 40.6 & 56.5 & 69.4 & 49.2 & 45.0 & 49.5 & 38.0 & 43.1 & 47.7\\
VideoPose3D \cite{pavllo20193d} & CVPR'19 & 36.0 & 38.7 &38.0 &41.7 &40.1 &45.9 &37.1 &35.4 &46.8 &53.4 &41.4 &36.9 &43.1 &30.3 & 34.8 & 40.0 \\
STGCN \cite{cai2019exploiting} & ICCV'19 & 36.8 & 38.7 & 38.2 & 41.7 & 40.7 & 46.8 & 37.9 & 35.6 & 47.6 & 51.7 & 41.3 & 36.8 & 42.7 & 31.0 & 34.7 & 40.2 \\
SRNet \cite{zeng2020srnet} & ECCV'20 &35.8 &39.2& 36.6 & \textcolor{red}{36.9} & 39.8 &45.1 &38.4& 36.9 & 47.7 & 54.4 & \textcolor{red}{38.6} & 36.3& \textcolor{red}{39.4} &30.3 &35.4& 39.4\\
LCN \cite{ci2019optimizing}  & ICCV'21 &36.9 &41.6 &38.0 &41.0& 41.9 &51.1 &38.2 &37.6 &49.1 &62.1 &43.1 &39.9 &43.5 &32.2 &37.0 & 42.2 \\
% Liu \textit{et al.} \cite{liu2020comprehensive} &35.9&40.0&38.0 &41.5 &42.5 &51.4 &37.8 &36.0 &48.6& 56.6& 41.8 &38.3 &42.7 &31.7 &36.2 &41.2\\
% MGCN \cite{zou2021modulated} &  ICCV'21 $\S$ & 35.7 & 38.6 & 36.3 & 40.5 & 39.2 & 44.5 & 37.0 & 35.4 & 46.4 & 51.2 & 40.5 & 35.6 & 41.7 & 30.7 & 33.9 & 39.1 \\
MLP-JCG \cite{tang2023mlp} & TMM'23 & \textcolor{red}{33.7} & \textcolor{blue}{37.4} & 37.3 & 39.6 & 39.8 & 47.1 & \textcolor{red}{33.7} & \textcolor{red}{33.8} & \textcolor{blue}{45.7} & 60.5 & \textcolor{blue}{39.7} & 37.7& \textcolor{blue}{40.1} & \textcolor{blue}{30.1} & \textcolor{blue}{33.8} & 39.3 \\
GKONet \cite{hu2023geometric} & TCSVT'23 & 35.4 &38.8 & \textcolor{blue}{35.9} &40.4 & 39.6 & 44.0 &36.7 &35.4& 46.8 &53.7 & 40.9 & 36.6 & 42.0 & 30.6 & 33.9 & 39.4 \\
ZEDO \cite{jiang2024back} &  WACV'24 & - & - & - & - & - & - & - & - & - & - & - & - & - & - & - & 42.1\\
\toprule
\multicolumn{2}{c|}{\textbf{Probabilistic Method}} & Dire. & Disc. & Eat & Greet & Phone & Photo & Pose & Purch. & Sit & SitD. & Smoke & Wait & WalkD. & Walk & WalkT. & Avg $\downarrow$  \\
\midrule
CVAE ($N$=200) \cite{sharma2019monocular} & ICCV'19 & 35.3 & \textcolor{red}{35.9} & 45.8 & 42.0 & 40.9 & 52.6 & 36.9 & 35.8 & \textcolor{red}{43.5} & 51.9 & 44.3 & 38.8 & 45.5 & \textcolor{red}{29.4} & 34.3 & 40.9 \\
GAN ($N$=10) \cite{li2020weakly} & BMVC'20 & 41.4 & 44.3 & 44.6 & 50.2 & 49.3 & 51.8 & 40.1 & 46.2 & 57.7 & 72.7 & 48.7 & 45.4 & 49.6 & 43.8 & 43.3 & 48.7 \\
GraphMDN ($N$=5) \cite{oikarinen2021graphmdn} & IJCNN'21 & 39.7 & 43.4 & 44.0 & 46.2 & 48.8 & 54.5 & 39.4 & 41.1 & 55.0 & 69.0 & 48.0 & 43.7 & 49.6 & 38.4 & 42.4 & 46.9 \\
NF ($N$=1) \cite{wehrbein2021probabilistic} & ICCV'21 & 37.8 & 41.7 & 42.1 & 41.8 & 46.5 & 50.2 & 38.0 & 39.2 & 51.7 & 61.8 & 45.4 & 42.6 & 45.7 & 33.7 & 38.5 & 43.8 \\ 
DiffPose ($N$=5) \cite{gong2023diffpose} & CVPR'23 & \textcolor{blue}{33.9} & 38.2 & 36.0 & \textcolor{blue}{39.2} & 40.2 & 46.5 & \textcolor{blue}{35.8} & 34.8 & 48.0 & 52.5 & 41.2 & 36.5 & 40.9 & 30.3 & \textcolor{blue}{33.8} & 39.2 \\
ProPose ($N$=1) \cite{han2025propose} & AAAI'25 & - & - & - & - & - & - & - & - & - & - & - & - & - & - & - & 40.4 \\
\midrule
\rowcolor{grey!10}  FMPose3D ($N$=2)  (Ours) & & 35.4 & 38.3 & 36.0 & 39.8 & \textcolor{blue}{39.2} & \textcolor{blue}{43.5} & 36.5 & 34.7 & 46.3 & \textcolor{blue}{48.4} & 40.4 & \textcolor{blue}{35.9} & 41.0 & 31.0 & 34.2 & \textcolor{blue}{38.7} \\
\rowcolor{grey!10}  FMPose3D ($N$=40) (Ours) & & 35.0 & 37.7 & \textcolor{red}{35.7} & 39.4 & \textcolor{red}{38.8} & \textcolor{red}{43.0} & 36.1 & \textcolor{blue}{34.2} & \textcolor{blue}{45.7} & \textcolor{red}{48.1} & 40.1 & \textcolor{red}{35.5} & 40.6 & 30.6 & \textcolor{red}{33.7} &  \textcolor{red}{38.3} \\
% pretrained_model/FM_GAMLP_noisePose_layers5_1GCNParallelAttnMLP_attnD_0.2_projD_0.25_lr1e-3_decay0.98_lr_decay_large_e5_0.8_B256_20250916_1953/exp_augH_s3_Top10_exp_temp0.005_h1,2,3,4,5,6,7,8,9,10,11,12,13,14,15,16,17,18,19,20_test_results_20251005_155902
% FMPose3D ($N$=10) (Ours) & 34.9 & 37.6 & 35.6 & 39.3 & 38.7 & 42.9 & 36.0 & 34.1 & 45.6 & 47.8 & 39.9 & 35.3 & 40.5 & 30.5 & 33.6 & \textbf{38.1} \\
\bottomrule
\end{tabular}
}
\label{tab:human3.6M_cpn_p2}
\end{table*}

\begin{figure*}[t]
\centering
\centerline{\includegraphics[width=0.95\linewidth]{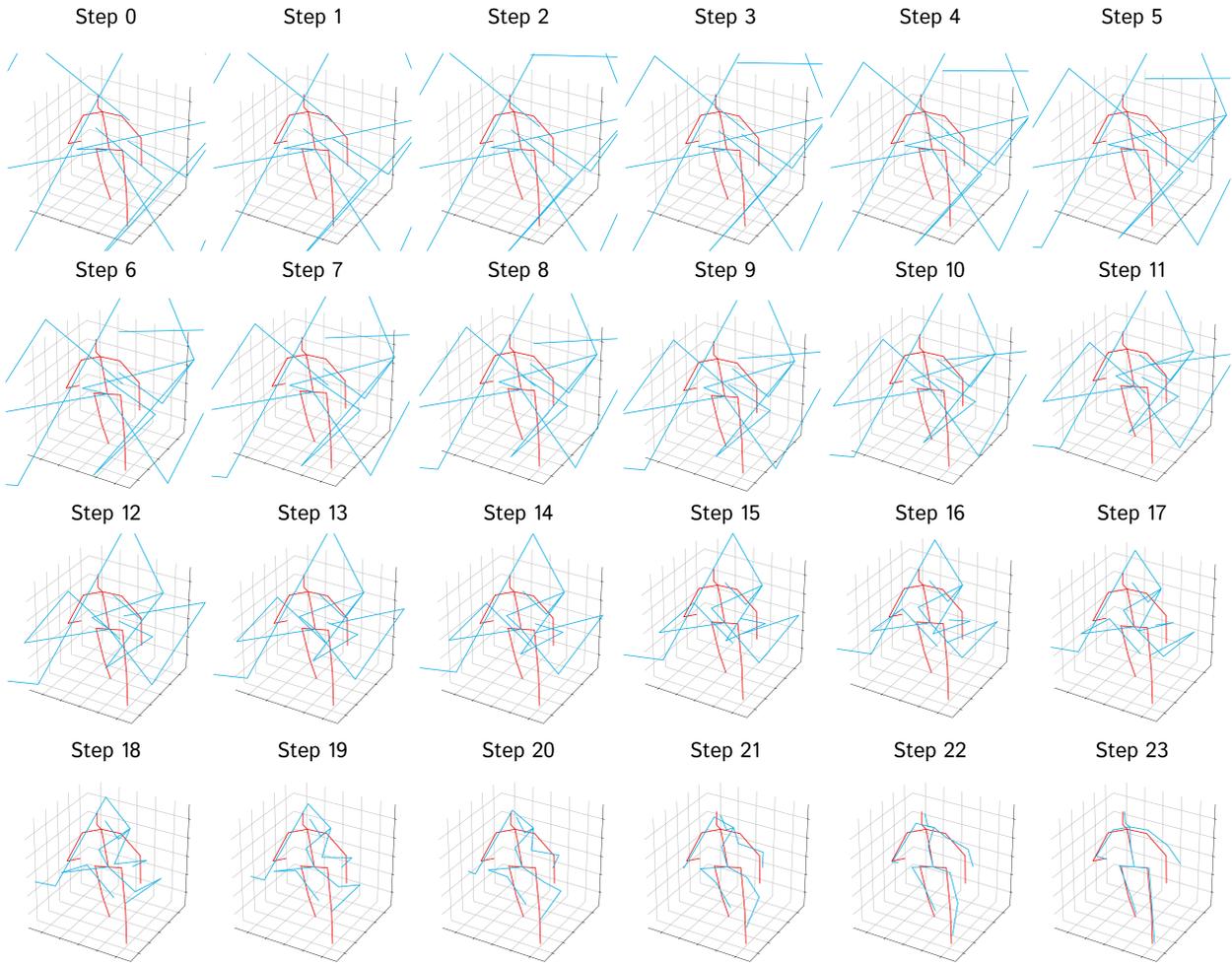}}
\caption
{Visualization of intermediate 3D pose predictions during inference with $S = 23$ integration steps. The blue pose represents the predicted results, while the red pose represents the ground truth.
}
\label{fig:diff_S}
\end{figure*}

\section{Additional Quantitative Results}
\label{sec:supp_h36m_p2}
\noindent \textbf{Human3.6M.}
Table~\ref{tab:human3.6M_cpn_p2} reports the P-MPJPE results on Human3.6M, comparing our FMPose3D with prior state-of-the-art deterministic methods (top) and probabilistic methods (bottom).
Our baseline model achieves 38.7 mm. When we increase the number of hypotheses and applying our RPEA module, the error is further reduced to 38.3 mm, demonstrating the effectiveness of multi-hypothesis modeling.

% \section{Additional Ablation Studies}
% \noindent \textbf{Impact of Integration Steps.}

\section{Intermediate Integration States}
\label{sec:suppl_vis_diff_S}
To better understand how the learned velocity field transports a noise sample toward the target 3D pose, we select one example, set the number of integration steps to $S = 23$, and visualize the intermediate predictions along the trajectory. As illustrated in Figure~\ref{fig:diff_S}, the poses start from random noise and progressively become more structured, gradually converging to a plausible human configuration that closely matches the final target pose.

\section{Impact of Integration Steps}
\label{sec:suppl_impact_steps}
During inference, our FMPose3D generates 3D poses by solving the underlying ODE from noise, conditioned on 2D inputs, using $S$ integration steps. Figure~\ref{fig:impact_steps} reports MPJPE and P-MPJPE on the Human3.6M dataset with 2D poses detected by CPN~\cite{chen2018cascaded} for different choices of $S$. 
With a single integration step, the errors of both metrics are relatively large. As $S$ increases, the errors first decrease, reach the minimum at $S = 4$, and then gradually increase. The results indicate that steps in the range $S \in \{3,4,5\}$ yield comparable accuracy, whereas larger $S$ does not provide additional gains. To strike a balance between estimation accuracy and computational efficiency, we set $S = 3$ in all experiments.

\begin{figure}[t]
    \centering
    \includegraphics[width=0.95\linewidth]{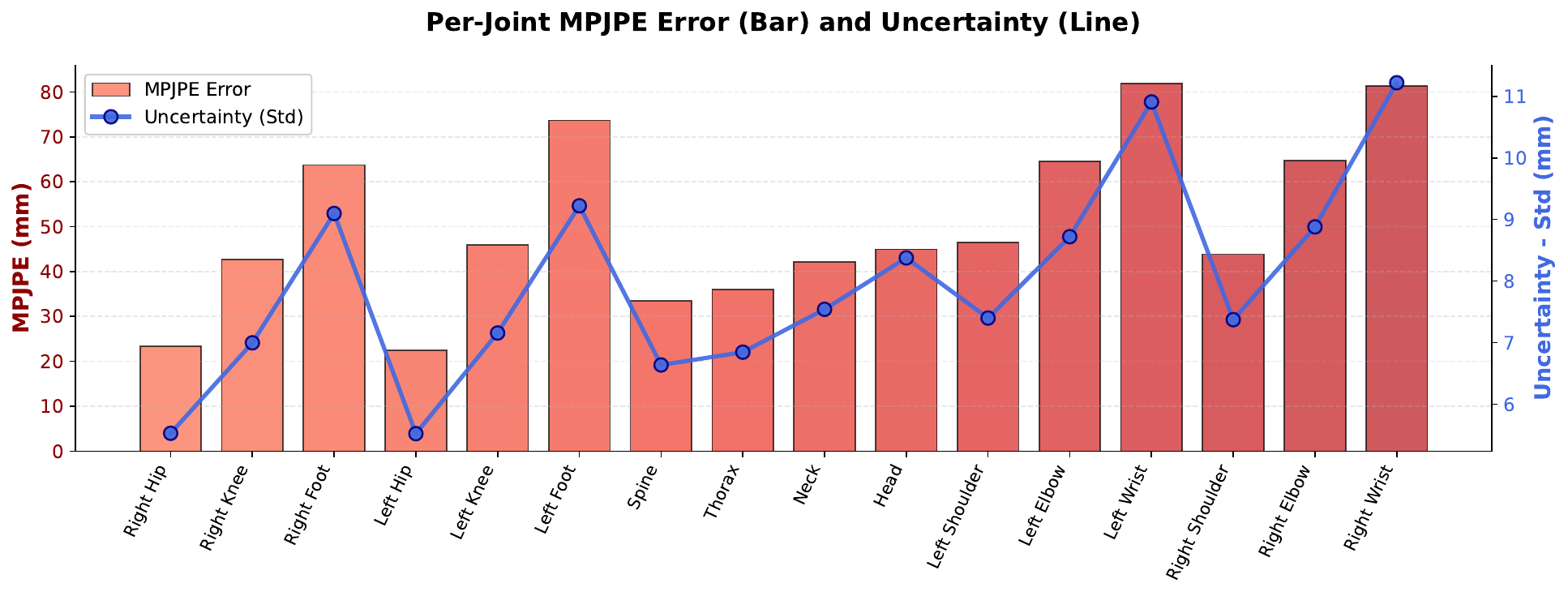}
    \caption{ Per-joint uncertainty versus per-joint error on Human3.6M. Uncertainty is measured as the standard deviation (Std) across hypotheses, and error is measured by MPJPE (mm).}
    \label{fig:uncertainty_per_joint}
\end{figure}

\begin{figure}[t]
    \centering
    \includegraphics[width=0.95\linewidth]{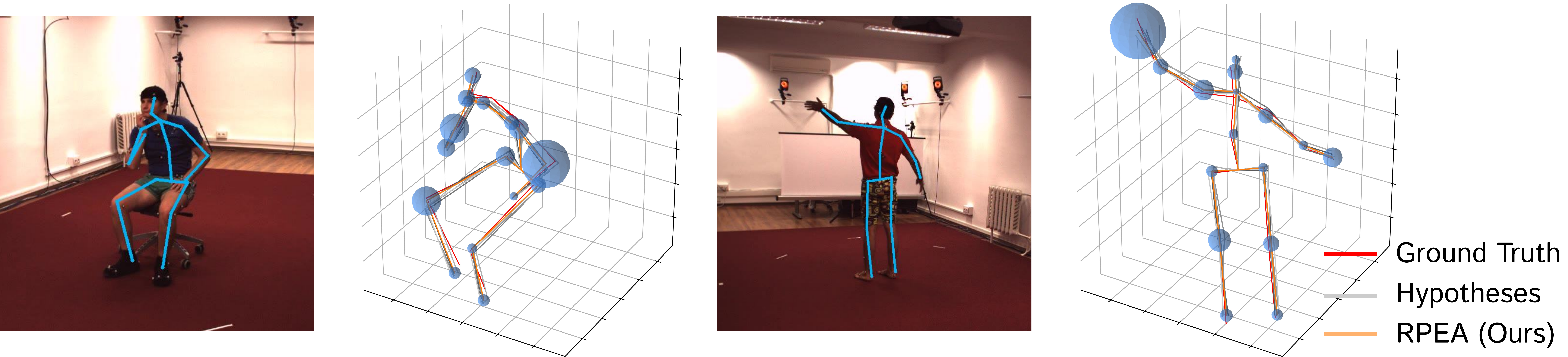}
    \caption{Uncertainty visualization. Left: highest uncertainty at the left elbow. Right: highest uncertainty at the left wrist.}
    \label{fig:vis_uncertainty}
\end{figure}

% \begin{figure}[t]
%     \centering
%     \includegraphics[width=0.95\linewidth]{figures/rpea_pose_level.pdf}
%     \caption{
%     Comparison of aggregation strategies on Human3.6M. We additionally report the performance of pose-level RPEA compared to Figure \ref{fig:aggregation}.
% The top plot reports MPJPE, while the bottom plot shows P-MPJPE.
%     }
%     \label{fig:aggregation_rpea_poselevel}
% \end{figure}

% \section{Pose Level Aggregation with RPEA}
% P-MPJPE reflects pose structure up to rigid alignment and is less sensitive to absolute 3D error, which is measured by MPJPE. In the main paper, we apply RPEA at the \emph{joint} level, optimizing each joint independently and thus ignoring global skeletal consistency; this yields a large MPJPE gain but only a slight P-MPJPE improvement. RPEA can also be applied at the \emph{pose} level: Figure \ref{fig:aggregation_rpea_poselevel} additionally reports this variant and shows that it achieves the best P-MPJPE, at the cost of slightly worse MPJPE.

\section{Uncertainty Estimation}
\label{sec:supp_uncertainty}
Our multi-hypothesis predictions also enable uncertainty estimation.
On the Human3.6M test set, we generate $N{=}40$ hypotheses per input and estimate uncertainty as the per-joint standard deviation across these hypotheses.
As shown in Figure \ref{fig:uncertainty_per_joint}, the average uncertainty is positively correlated with the per-joint error, supporting the validity of this uncertainty measure. Such single-view uncertainty can serve as a confidence signal for unsupervised multi-view fusion and as an intermediate representation to guide mesh reconstruction.

For visualization, we represent uncertainty using spheres centered at the predicted joint locations, with radii proportional to the per-joint variance.
Larger spheres indicate higher uncertainty, as illustrated in Figure ~\ref{fig:vis_uncertainty}.

\section{Results on an Additional Benchmark: 3DPW}
\label{sec:supp_3dpw}
In the main paper, we follow common practice in monocular 3D pose estimation and primarily report results on Human3.6M and MPI-INF-3DHP for fair comparison with prior work.
We now additionally evaluate on 3DPW~\cite{pw3d} with our Human3.6M pretrained model, the results are reported in Table~\ref{table:3dpw}. 
Since 3DPW is more commonly used in human mesh reconstruction, the methods listed in Table~\ref{table:3dpw} are primarily mesh-based approaches, and thus the comparison should be interpreted with this difference in mind.
Our model achieves the best zero-shot performance, comparable even to models pretrained on this dataset.

\begin{table}[t]
    \centering
    \caption{Results on 3DPW. Top: methods trained on 3DPW. Bottom: methods without 3DPW training (zero-shot evaluation).}
    \setlength\tabcolsep{4.5mm}
    \footnotesize
    \begin{tabular}{l|cc}
    \toprule
    Method & P-MPJPE & MPJPE \\
    \midrule
    HMR2.0a \cite{goel2023humans} (ICCV23) & 44.5 & 70.0 \\
    Multi-HMR \cite{baradel2024multi} (ECCV24) & 41.7 & 61.4 \\
    \midrule
    AdaptPose \cite{gholami2022adaptpose} (CVPR22) & 46.5 & 81.2 \\
    PMCE \cite{you2023co} (ICCV23) & 52.3 & 81.6 \\
    ScoreHMR \cite{stathopoulos2024score} (CVPR24) & 50.5 & - \\
    \rowcolor{blue!15} FMPose3D ($N$=40) (Ours) & 42.4 & 70.9 \\
    \bottomrule
    \end{tabular}
\label{table:3dpw}
\end{table}

\begin{table}[t]
    \centering
    \caption{Effect of training set size. For each dataset, the model is trained from scratch on randomly subsampled training subsets of 10\%, 20\%, 40\%, and 80\%. We run each setting three times with different random seeds and report the mean and the standard deviation. Full training and test set sizes in frames (train, test): Human3.6M (3,119k, 543k), Animal3D (3k, 0.3k), CtrlAni3D (8k, 1.4k).} 
    \setlength{\tabcolsep}{1.9mm}
    \footnotesize
    \begin{tabular}{c|cc|c|c}
    \toprule
    \multirow{2}{*}{Data (\%)}
      & \multicolumn{2}{c|}{Human3.6M} 
      & \multicolumn{1}{c|}{Animal3D} 
      & \multicolumn{1}{c}{CtrlAni3D} \\
    & P-MPJPE & MPJPE & P-MPJPE & P-MPJPE \\
    \midrule
    10 & $43.5_{\pm 0.85}$ & $53.9_{\pm 0.71}$  & $120.7_{\pm 1.88}$ & $83.5_{\pm 0.95}$ \\
    20 & $41.8_{\pm 0.62}$ & $52.2_{\pm 0.64}$ & $98.3_{\pm 0.89}$ & $69.5_{\pm 0.30}$ \\
    40 & $40.4_{\pm 0.33}$ & $50.7_{\pm 0.32}$ & $81.6_{\pm 0.28}$ & $57.8_{\pm 0.24}$ \\
    80 & $39.3_{\pm 0.15}$ & $49.7_{\pm 0.09}$ & $67.2_{\pm 1.33}$ & $47.9_{\pm 1.02}$ \\
    \bottomrule
    \end{tabular}
    \label{table:data_scaling_3datasets}
\end{table}

\begin{figure*}[htbp]
\centering
\centerline{\includegraphics[width=1.0\linewidth]{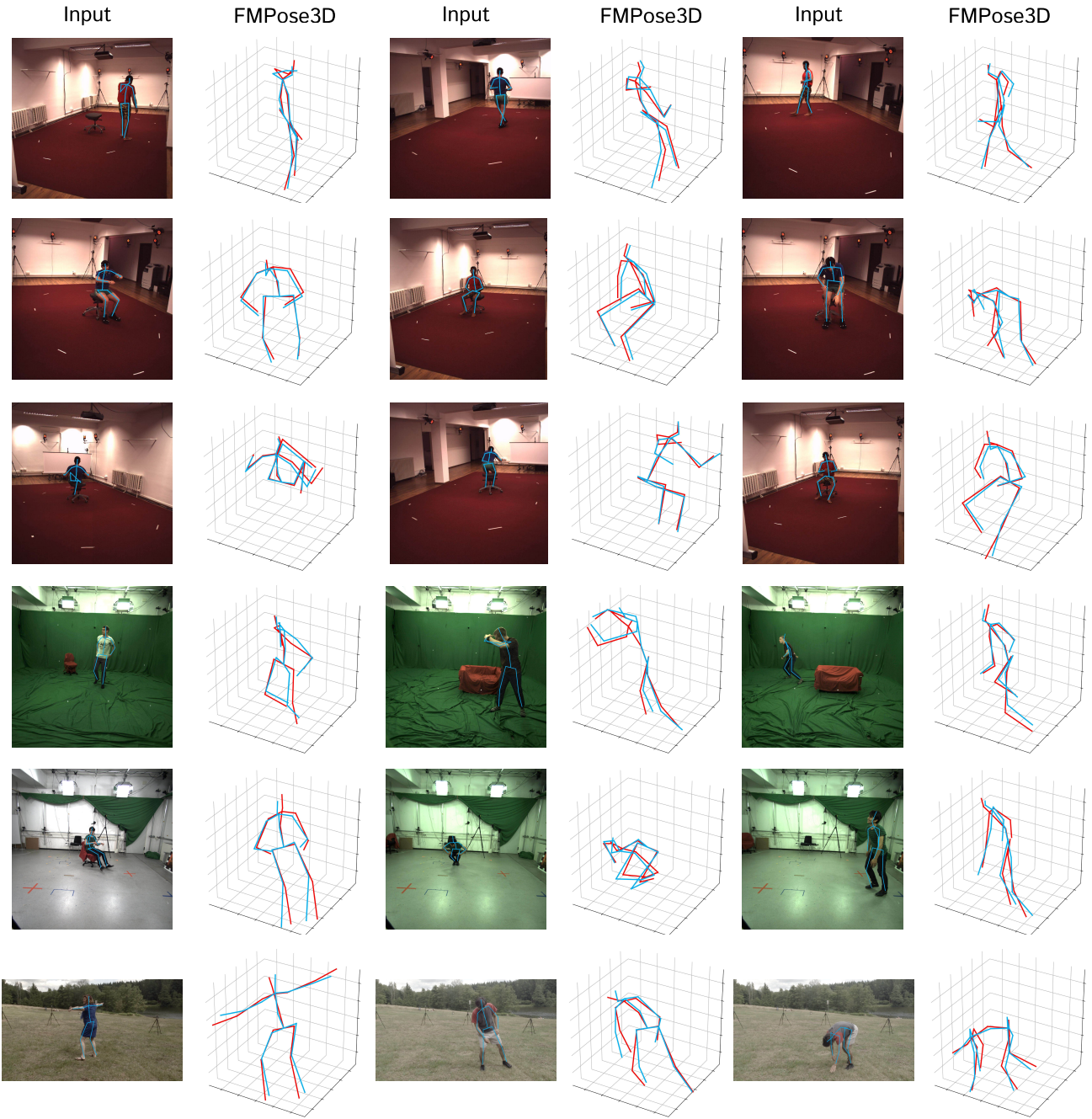}}
\caption
{
Qualitative results on Human3.6M (top three rows) and MPI-INF-3DHP (bottom three rows). The blue pose represents the predicted results, while the red pose represents the ground truth.}
\label{fig:humans}
\end{figure*}

\begin{figure*}[htbp]
\centering
\centerline{\includegraphics[width=0.85\linewidth]{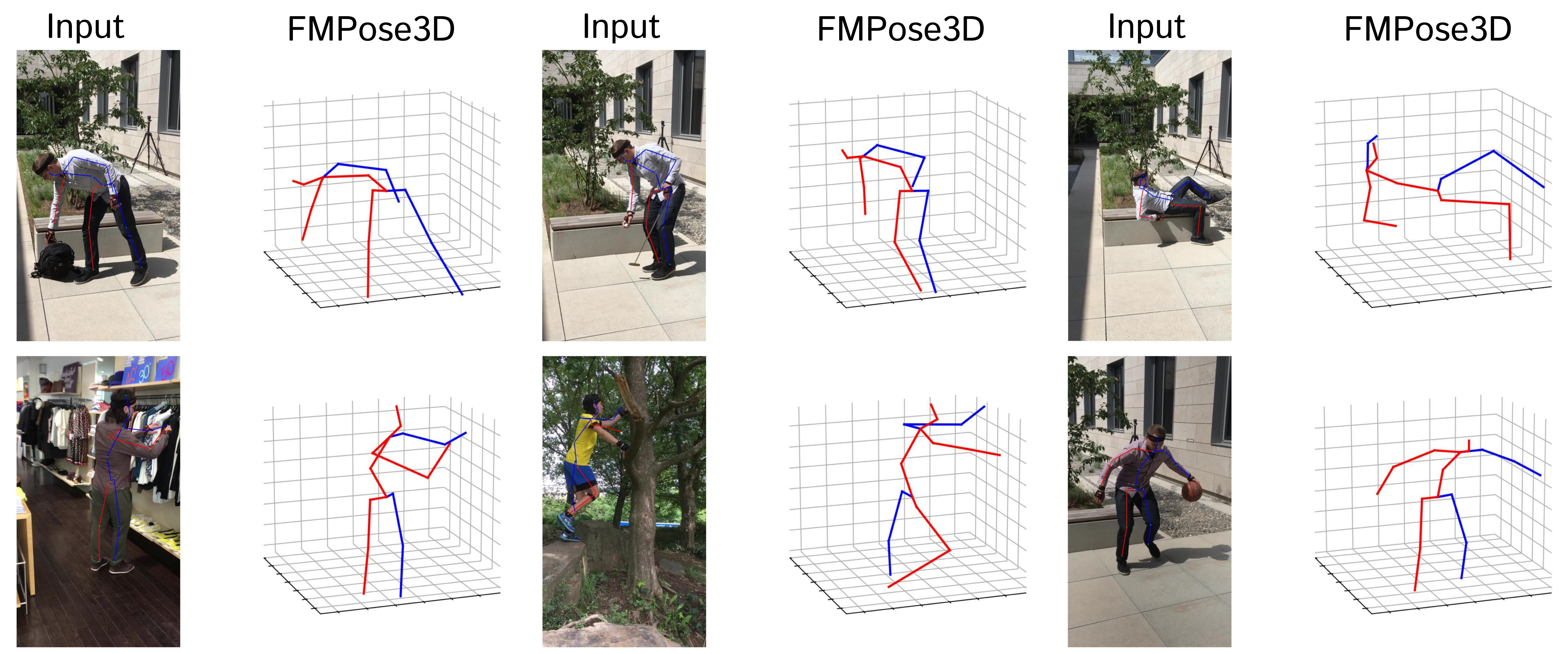}}
\caption
{Qualitative results on 3DPW. The 2D pose is detected by HRNet \cite{wang2020deep}.
}
\label{fig:vis_3dpw}
\end{figure*}

\begin{figure*}[t]
\centering
\centerline{\includegraphics[width=0.75\linewidth]{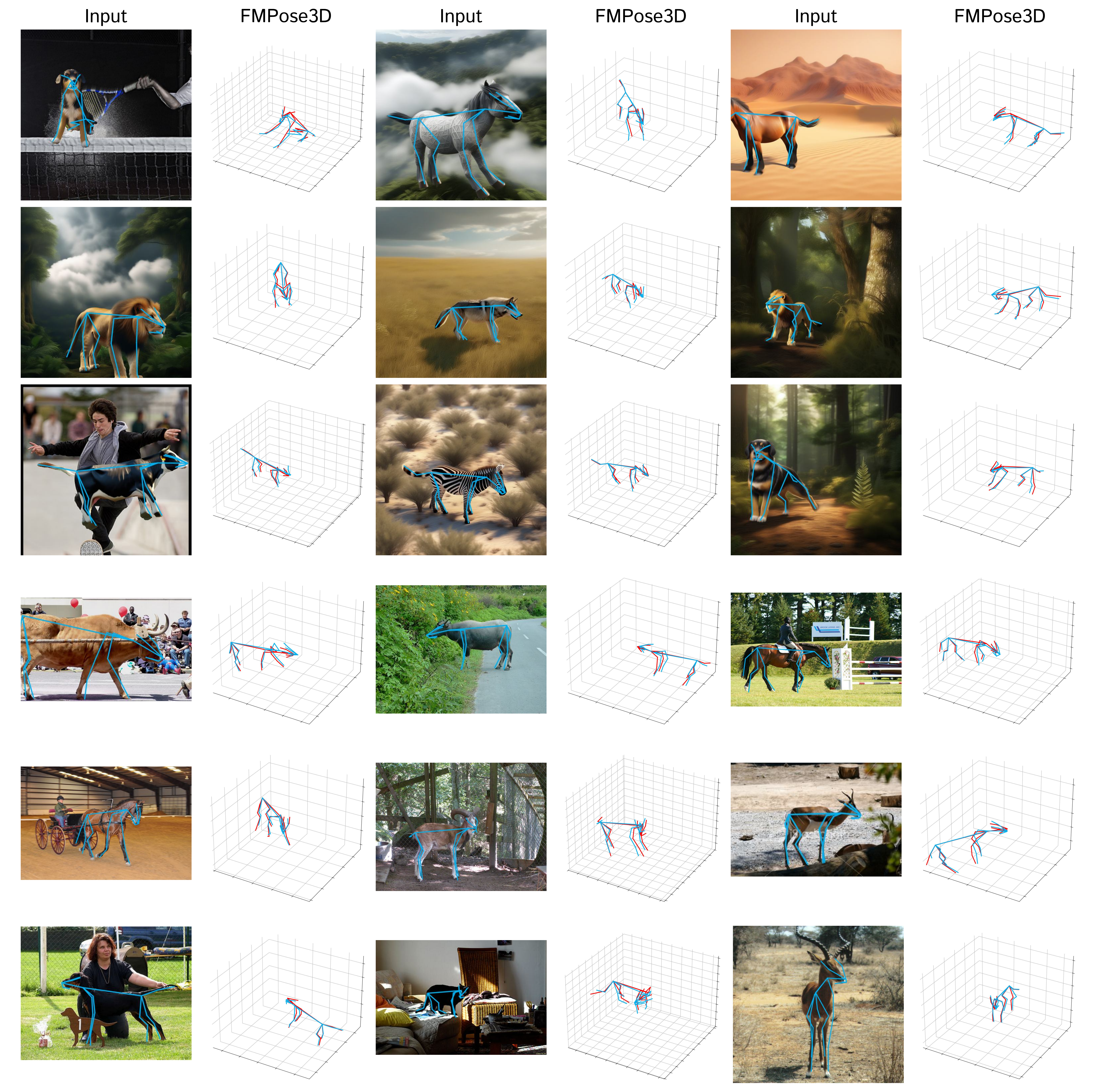}}
\caption
{Qualitative results on CtrlAni3D (top three rows) and Animal3D (bottom three rows). The blue pose represents the predicted results, while the red pose represents the ground truth.
}
\label{fig:animals}
\end{figure*}

\section{Effect of training set size}
\label{sec:supp_effect_training_set_size}
To evaluate the sensitivity to training data size, we train our model from scratch on randomly subsampled subsets comprising 10\%, 20\%, 40\%, and 80\% of the training set. For each fraction, we run three independent trainings with different random seeds and report the mean performance on the full test set. Table \ref{table:data_scaling_3datasets} shows consistent improvements as the training set size increases across all datasets. The gains are modest on Human3.6M, but much larger on Animal3D and CtrlAni3D, as these animal datasets are relatively small and thus more sensitive to reduced training data.

\section{Additional Qualitative Results}
\label{sec:suppl_qualitative}
Figure~\ref{fig:humans} presents qualitative results of the proposed FMPose3D on the Human3.6M and MPI-INF-3DHP datasets. The model is trained solely on Human3.6M.
Human3.6M consists of indoor scenes (top three rows), while the MPI-INF-3DHP test set includes three scenarios: studio with green screen (GS, fourth row), studio without green screen (noGS, fifth row), and outdoor scenes (Outdoor, sixth row).

% 3DPW~\cite{pw3d} is a challenging in-the-wild dataset. 
To further assess the generalization ability of our model to outdoor scenarios, we evaluate the model pre-trained on Human3.6M using samples from the 3DPW ~\cite{pw3d} dataset. Figure~\ref{fig:vis_3dpw} presents the qualitative results. The 2D poses are obtained using HRNet~\cite{wang2020deep}. The results indicate that our model generalizes well to unconstrained in-the-wild environments.

Figure~\ref{fig:animals} further shows qualitative results on animal datasets, including synthetic samples from CtrlAni3D (top three rows) and real-world samples from Animal3D (bottom three rows). Across these challenging cases, our approach consistently produces reliable and anatomically plausible 3D pose predictions.
\end{document}